\documentclass{article}

\usepackage{microtype}
\usepackage{graphicx}
\usepackage{subcaption}
\usepackage{booktabs} 

\usepackage{microtype}
\usepackage{graphicx}
\usepackage{booktabs} 
\usepackage{multirow}
\usepackage{epstopdf}
\usepackage{amssymb}
\usepackage{amsmath}
\usepackage{amsthm}             
\usepackage{wrapfig}

\usepackage{mathtools}

\usepackage{colortbl}
\usepackage[table]{xcolor}

\usepackage{placeins}

\usepackage[most]{tcolorbox}
\usepackage{xcolor}

\definecolor{promptframe}{RGB}{60,60,60}
\definecolor{promptbg}{RGB}{248,248,248}

\newtcolorbox{promptbox}[1]{%
  enhanced,
  breakable,
  colback=promptbg,
  colframe=promptframe,
  boxrule=0.6pt,
  arc=2pt,
  left=6pt,right=6pt,top=6pt,bottom=6pt,
  fonttitle=\bfseries\small,   
  fontupper=\small,            
  colbacktitle=black,   
  coltitle=white,       
  title={#1},
  before skip=6pt plus 2pt minus 2pt,
  after skip=6pt plus 2pt minus 2pt,
}

\usepackage{hyperref}

\usepackage[accepted]{icml2026}

\usepackage{tikz}
\newcommand{\blackcircnum}[1]{%
  \tikz[baseline=(char.base)]{
    \node[shape=circle, fill=black, text=white, inner sep=0.1pt, minimum size=1pt] (char) {#1};
  }%
}

\usepackage[capitalize,noabbrev]{cleveref}

\theoremstyle{plain}
\newtheorem{theorem}{Theorem}[section]

\theoremstyle{definition}

\newtheorem{assumption}[theorem]{Assumption}
\theoremstyle{remark}

\usepackage[textsize=tiny]{todonotes}

\icmltitlerunning{Rethinking Efficient Graph Coarsening via a Non-Selfishness Principle}

\begin{document}

\twocolumn[
  \icmltitle{Rethinking Efficient Graph Coarsening via a \textit{Non-Selfishness} Principle}



  \icmlsetsymbol{equal}{*}

  \begin{icmlauthorlist}
    \icmlauthor{Xu Bai}{yyy}
    \icmlauthor{Bin Lu}{equal,yyy}
    \icmlauthor{Kun Zhang}{qqq}
    \icmlauthor{Shengbo Chen}{uuu}
    \icmlauthor{Xinbing Wang}{yyy}
    \icmlauthor{Chenghu Zhou}{iii}
    \icmlauthor{Meng Jin}{equal,rrr}

  \end{icmlauthorlist}

  \icmlaffiliation{yyy}{School of Information Science and Electronic Engineering, Shanghai Jiao Tong University, Shanghai, China}
  \icmlaffiliation{rrr}{School of Artificial Intelligence, Shanghai Jiao Tong University, Shanghai, China}
  \icmlaffiliation{qqq}{School of Environment Science and Engineering, Shanghai Jiao Tong University, Shanghai, China}
  \icmlaffiliation{uuu}{School of Artificial Intelligence, Nanchang University, Nanchang, China}
  \icmlaffiliation{iii}{Institute of Geographic Sciences and Natural Resources Research, Chinese Academy of Sciences, Beijing, China}

  \icmlcorrespondingauthor{Bin Lu}{robinlu1209@sjtu.edu.cn}
  \icmlcorrespondingauthor{Meng Jin}{jinm@sjtu.edu.cn}

  \icmlkeywords{Machine Learning, ICML}

  \vskip 0.3in
]



\printAffiliationsAndNotice{}  

\begin{abstract}
Graph coarsening is a graph dimensionality reduction technique that aims to construct a smaller and more tractable graph while preserving the essential structural and semantic properties of the original graph. However, most existing methods rely on pair-wise similarity matching, where each node independently searches for its best partner based on global information. This \textit{selfishness} matching paradigm incurs substantial computational and memory overhead.
To address this problem, we shift to a \textit{non-selfishness} principle that prioritizes the collective interference of neighborhood in coarsening, and propose an efficient method named \texttt{NOPE}, which achieves linear memory consumption and near-linear computational complexity in the number of nodes.
Furthermore, we derive a faster variant \texttt{NOPE}$^{*}$, which reduces $\mathcal{O}(\Delta\cdot d)$ interference evaluation to $\mathcal{O}(d)$ based on the local isotropy assumption, and consequently alleviates the computational bottleneck for high-degree nodes. Experimental results show that \texttt{NOPE}$^{*}$ achieves $1.8–10\times$ speedup over \texttt{NOPE} and surpass almost all baselines with 1-3 orders of magnitude acceleration. Meanwhile, learning on coarsened graphs yields comparable performance to original graphs, and can even show superior performance over LLM-based graph reasoning owing to compact graph information. 
The code can be available at https://github.com/dazonglian/NOPE-main.
\end{abstract}

\vspace{-6mm}

\section{Introduction}

Graph-structured data is ubiquitous in real-world applications, and graph learning methods—such as graph neural networks (GNNs)~\cite{gnn_graph} and large language models (LLMs) for graph reasoning~\cite{llm_graph, zhao2026inductive,wu2026sogk}—have achieved promising performance~\cite{lu2024oxygenerator}. However, the increasing scale and complexity of modern graphs impose substantial constraints for hardware and resources, limiting the practical applicability of existing approaches~\cite{fgc_icml, bai2024dual, bai2025dual}. Graph coarsening addresses this challenge through graph dimensionality reduction~\cite{gc_inf_sci, semi_gc}. By aggregating nodes and edges into a smaller, more tractable graph, coarsening aims to preserve the essential structural and semantic properties of the original graph, enabling efficient analysis and downstream learning with minimal loss of information fidelity~\cite{gc, a-cm, gc_cbm}.

Most existing graph coarsening methods adopt a pair-wise similarity matching paradigm, where nodes independently select merge partners based on similarity, e.g., FGC~\cite{fgc_jmlr} merges nodes with high similarity by selecting pairs that minimize spectral loss, whereas A-CM~\cite{a-cm} conducts similarity-driven merging using representations obtained from a global graph convolution. These strategies are inherently \textbf{\textit{selfish}}, as it considers only pair-wise affinity while neglecting the impact of merging on surrounding neighborhoods, which can distort neighborhood-level semantics, particularly in text-attributed graphs. Moreover, accurate similarity evaluation and partner search typically rely on global information, leading to substantial computational and memory overhead. These limitations motivate the need for a graph coarsening principle that moves beyond \textit{selfish} pair-wise decisions while remaining both scalable and semantically faithful.

To this end, we propose a \textbf{\emph{non-selfishness principle}} for graph coarsening, which prioritizes merges that minimize disturbance to local neighborhoods rather than relying solely on pair-wise similarity. 
To formalize this principle, we propose the neighborhood interference index ($\mathcal{I}$), a metric that quantifies the semantic deviation induced by a candidate merge. 
Guided by this index, we also develop \textbf{\underline{No}}n-selfishness \textbf{\underline{P}}rinciple Graph Coars\textbf{\underline{E}}ning (\texttt{NOPE}), a greedy interference-aware coarsening algorithm that achieves linear memory usage and near-linear time complexity. To further improve scalability as supernode degrees increase during coarsening, we derive a faster variant, \texttt{NOPE}$^*$. It leverages a local isotropy assumption to reduce the per-merge evaluation cost from $\mathcal{O}(\Delta \cdot d)$ to $\mathcal{O}(d)$, where $\Delta$ denotes the dynamic average degree and $d$ denotes the feature dimension.

In summary, this work makes the following contributions.
\begin{enumerate}
    \item Rethinking existing graph coarsening methods, we are the first to shift the paradigm to a \textit{non-selfishness} principle, explicitly considering the minimization of neighborhood-level interference. Hereby, we propose \texttt{NOPE}, achieving a linear memory consumption and near-linear time complexity.
    \item To further eliminating the increasing cost of degree-dependent interference quantification, we theoretically derive a faster variant \texttt{NOPE$^{*}$}, reducing the per-merge complexity from $\mathcal{O}(\Delta\cdot d)$ to $\mathcal{O}(d)$.
    \item Extensive experiments demonstrate that \texttt{NOPE$^{*}$} achieves 1–3 orders of magnitude speedup over baselines while preserving or improving performance over both GNN- and LLM-based methods.
\end{enumerate}

\section{Related Work}

\textbf{Structure Graph Coarsening}. The primary goals of structure graph coarsening algorithms are to reduce the computational burden of graph algorithms or to minimize graph storage~\cite{compressgraph, Laconic, mags, 2023_gc_str}. Their coarsening rules mainly focus on structural properties of graphs, such as preserving topological similarity between the coarsened graph and the original graph~\cite{query_gc, gc_weight_graph}, or merging structurally similar nodes to reduce the number of edges and achieve compact graph representations~\cite{lsh_graph_summarization, graph_sum_survey, grass}. However, these methods operate solely at the structural level and ignore node features or textual attributes, making them unsuitable for our setting.

\textbf{Feature Graph Coarsening}. For feature graph coarsening, FGC~\cite{fgc_jmlr} first introduce $\epsilon-similarity$ to quantify feature consistency between the coarsened graph and the original graph, and formulate the problem as an optimization objective. Building on this idea, UGC~\cite{ugc_nips} improve efficiency by adopting a hash-based grouping strategy to merge similar nodes, and further propose AH-UGC~\cite{ah-ugc}, which enables continuous graph coarsening. Moreover, several GNN-based methods accelerate GNN computation through graph coarsening. MPG~\cite{gc-mpg_nips}, A-CM~\cite{a-cm}, and CoCoA~\cite{cocoa} design coarsening criteria aligned with GNN-specific properties, such as message propagation similarity, to guide the coarsening process.

\textbf{Graph Condensation vs. Feature Graph Coarsening}. 
It should be noted that some graph condensation methods~\citep{gc_survey, gc_ntk, gc}, although also aimed at enhancing the scalability of GNNs, do not strictly perform graph coarsening in a technical sense. In summary, the following issues distinguish graph condensation from graph coarsening: 1. \textbf{Lack of universality}. These methods optimize objectives tailored to specific GNN models, which limits their applicability to downstream models and often produces condensed graphs that do not preserve essential graph structures and features. 2. \textbf{Black-box compression}. The condensed graphs are typically learned without explicit node or edge correspondences to the original graphs, resulting in limited interpretability.
In contrast, graph coarsening focuses on preserving the structural properties of the graph itself, yielding higher universality across downstream models and maintaining clear mappings between the original and coarsened graphs.

\section{Background}
\subsection{Notations}
Denote a text-attributed graph as $\mathcal{G}=\left(\mathcal{V},\mathcal{E},\mathcal{X},\mathcal{R}\right)$, where $\mathcal{V}$
denotes the node set, $\mathcal{E}$ is the edge set, $\mathcal{X}\in \mathbb{R}^{n\times d}(n=\left|\mathcal{V}\right|, d>0)$ denotes the feature matrix and $\mathcal{R}=\{R_1, R_2,...,R_{|\mathcal{V}|}\}$ represents the collection of raw text sequences associated with each node $v_i\in\mathcal{V}$. Each node $v_i$ is further assigned a ground-truth label $y_i\in\mathcal{Y}$. A complete summary of all notations used in this paper is provided in Appendix~\ref{tab:notation}.
\subsection{Problem Formulation}\label{prob_form}
Given an original text-attributed graph $\mathcal{G}=\left(\mathcal{V},\mathcal{E},\mathcal{X},\mathcal{R}\right)$, the goal of text-attributed graph coarsening is to construct a compressed graph $\mathcal{G}^c=\left(\mathcal{V}^c,\mathcal{E}^c,\mathcal{X}^c,\mathcal{R}^c\right)$ with a minimal number of supernodes $\mathcal{V}^c$ ($\left|\mathcal{V}^c\right|\ll\left|\mathcal{V}\right|,\left|\mathcal{V}^c\right|=n^c$) and corresponding edges $\mathcal{E}^c$, while each supernode $\mathcal{V}^c_i$ is related to concise text $\mathcal{R}^c_i$. The relationship between the nodes $\mathcal{V}$ in graph $\mathcal{G}$ and the supernodes $\mathcal{V}^c$ in $\mathcal{G}^c $ can be represented by a mapping matrix $\mathbf{C}$, which belongs to the following set
\begin{equation*}
    \mathcal{C} = \left\{ \mathbf{C}\in \{0,1\}^{n^c\times n} ~\middle|~ 
    \begin{aligned}
        & \langle \mathbf{C}_i, \mathbf{C}_j \rangle = 0, \forall i\ne j, \\
        & \|\mathbf{C}_{k}\|_0\geq 1, \forall k \in \{1,\dots,n^c\}
    \end{aligned}
    \right\}
\end{equation*}
where $\mathbf{C}_i$ and $\mathbf{C}_j$ represent $i$-th and $j$-th row of mapping matrix $\mathbf{C}$, $\langle \mathbf{C}_i, \mathbf{C}_j \rangle = 0$ means each original node must belong to and only belong to one supernode. $\|\mathbf{C}_{k}\|_0\geq 1$ denotes each supernode contains at least one original node. Additionally, we set $r= (1-\frac{n_c}{n})$ as the coarsening rate.
The node features $\mathcal{X}=\{\mathbf{x}_1, \mathbf{x}_2,...,\mathbf{x}_n\}$ are encoded from the raw text $\mathcal{R}$ via Sentence-BERT. For each supernode $v_i^c$, its feature representation $\mathbf{x}_i^c$ is obtained by averaging the embeddings of all its constituent nodes: $\mathbf{x}^c_i = \frac{1}{|v^c_i|} \sum_{j \in v^c_i} \mathbf{x}_j$.

\section{Methodology}
This section presents a graph coarsening framework grounded in a \textit{non-selfishness} principle. We define the neighborhood interference index ($\mathcal{I}$) to quantify the semantic disturbance induced by node aggregation. Based on it, we further propose a greedy coarsening algorithm, \texttt{NOPE}, which achieves near-linear time complexity and linear space complexity with respect to the number of nodes. To avoid exhaustive neighbor-wise similarity evaluations during coarsening, we introduce a faster variant, \texttt{NOPE}$^*$, which leverages a local isotropy assumption to enable a constant-time approximation of neighborhood similarity.

\subsection{Neighborhood Interference Index}
Here, we first introduce the neighborhood interference index ($\mathcal{I}$). Considering a coarsening operation where nodes $v_p^c$ and $v_q^c$ are merged to a new supernode $v_w^c$, the interference caused to their combined neighborhood is quantified as
\begin{equation*}
    \begin{split}
        \mathcal{I}_{pq} = \sum_{i\in \mathcal{N}_{p}\cup \mathcal{N}_q} |v_p^c|(s_{ip}-s_{iw})^2 + |v_q^c|(s_{iq}-s_{iw})^2.
    \end{split}
\end{equation*}
Here, the terms $(s_{ip} - s_{iw})^2$ and $(s_{iq} - s_{iw})^2$ quantify the information shift induced by merging. Specifically, they measure how much the original relationship between a neighbor $v_i^c$ and node $v_p^c$ (or $v_q^c$) deviates from its relationship with the resulting supernode $v_w^c$. These deviations are aggregated over the union of the neighbor sets and weighted by the corresponding node sizes $|v_p^c|$ and $|v_q^c|$, thereby reflecting the relative impact of each node involved in the merge. By explicitly penalizing neighborhood-level information shifts, this index discourages \textit{selfish} merging decisions that preserve only pairwise similarity while distorting the surrounding semantic structure, and instead favors \textit{non-selfish} merges that maintain neighborhood-level semantic stability. In this work, we adopt the dot product as the similarity measure, i.e., $s_{ij}=\langle \mathbf{x}_i, \mathbf{x}_j\rangle$.
Under this definition, $\mathcal{I}_{pq}$ can be written as
\vspace{-1mm}
\begin{equation}\label{i_simple}
    \begin{split}
        \mathcal{I}_{pq} = \sum_{i\in \mathcal{N}_{p}\cup \mathcal{N}_q} \frac{|v_p^c||v_q^c|}{|v_p^c|+|v_q^c|}(s_{ip}-s_{iq})^2.
    \end{split}
\end{equation}
The derivation process can be found in Appendix~\ref{emb_update} and~\ref{simp_pro_i}. 

\begin{figure}[h]
\begin{center}
\centerline{\includegraphics[width=1\columnwidth]{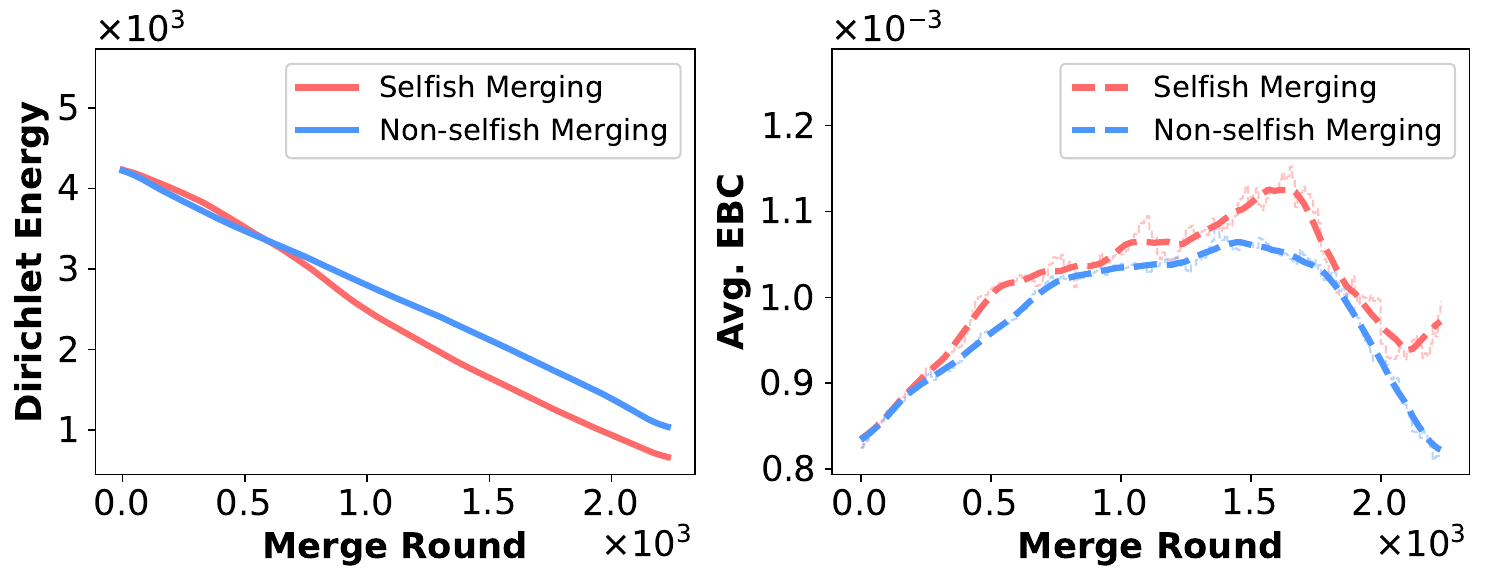}}
\caption{The changes of Dirichlet Energy and Average Edge Betweenness Centrality in the coarsened graphs based on \textit{selfishness} and \textit{non-selfishness} coarsening strategies.}
\label{reasonable_ana}
\end{center}
\end{figure}

To further illustrate the effectiveness of \textit{non-selfishness} strategies, we compare the typical \textit{selfish} strategy from two complementary perspectives: semantic preservation, measured by Dirichlet Energy, and structural balance, assessed via Average Edge Betweenness Centrality (Avg. EBC). For specific experimental settings, please refer to the  Appendix~\ref{reasonable_exp_design}. Figure~\ref{reasonable_ana} demonstrates that the \textit{non-selfishness} strategy consistently maintains higher Dirichlet Energy, directly confirming its effectiveness in preserving feature discriminability and preventing over-smoothing during coarsening. In contrast, the \textit{selfishness} baseline gradually eliminates feature differences. From the structural perspective, the smooth evolution of Avg. EBC under \textit{non-selfishness} merging indicates a balanced distribution of connectivity responsibility, preventing shortest-path flow from becoming overly concentrated on a small number of edges. In contrast, the sharp increase of Avg. EBC under \textit{selfish} merging reflects severe connectivity load concentration caused by improper disconnections. From a community perspective, it further intensifies flow pressure on inter-community edges, increasing the risk of community collapse.

\vspace{-2mm}

\subsection{\textit{Non-selfishness} Principle Graph Coarsening}\label{nope_section}
Motivated by our proposed neighborhood interference index $\mathcal{I}$, we design \texttt{NOPE}, an interference-aware greedy graph coarsening algorithm that iteratively contracts node pairs with minimal interference. To support efficient evaluation of $\mathcal{I}$ across iterations, \texttt{NOPE} precomputes the neighborhood interaction terms required for contraction decisions, and update and adjust them after each round of merging.

\subsubsection{\texttt{NOPE} Algorithm}
As outlined in Algorithm~\ref{alg:nope}, the execution flow is structured into four distinct phases:
\blackcircnum{1} \textbf{Cache Initialization}. We initialize a cache $\boldsymbol{\Sigma}$ to store, for each node, an aggregated measure of feature interactions with its neighborhood. This cache is designed to reuse intermediate results across iterations, thereby avoiding repeated computation of neighborhood statistics during the coarsening process.
\blackcircnum{2} \textbf{Priority Queue Construction}. Based on the initialized cache, we compute $\mathcal{I}$ for each edge in the graph. All edge candidates are then inserted into a priority queue $\mathcal{H}$, which is ordered by increasing values of $\mathcal{I}$. This queue defines a greedy contraction order that prioritizes merges expected to introduce minimal interference.
\blackcircnum{3} \textbf{Main Coarsening Loop}. At each iteration, we extract the edge $(u, v)$ with the smallest interference index from $\mathcal{H}$ and merge the corresponding nodes into a new supernode $w$. The embedding of $w$ is computed as a size-weighted average of the embeddings of $u$ and $v$. To ensure efficiency, obsolete edge candidates generated by previous merges are handled using a lazy removal strategy.
\blackcircnum{4} \textbf{Cache Updates}. After each merge operation, the cache $\boldsymbol{\Sigma}$ is updated incrementally to reflect the modified graph structure. Here we adjust the cached statistics of the affected neighbors to account for the replacement of nodes $u$ and $v$ by the supernode $w$. This incremental update maintains exactness while reducing computational overhead.

\subsubsection{\texttt{NOPE} Time and Space Complexity Analysis}\label{nope_complexity}

Assuming that the algorithm performs $\mathcal{O}(n)$ merge operations, \texttt{NOPE} has an overall time complexity of $\mathcal{O}(n \cdot \Delta^{2} \cdot d)$, where $\Delta$ denotes the dynamic average node degree during the coarsening process.
Since $\Delta \ll n$ in practical settings, the resulting time complexity is close to linear in the number of nodes.
The space complexity is dominated by the storage of node features and the graph structure, requiring $\mathcal{O}(n_{\max} d + |\mathcal{E}|)$ memory, where $n_{\max} = 2n$.
This leads to linear space complexity with respect to the input graph size and feature dimension, making the proposed method well suited for large-scale text-attributed graphs with high-dimensional features.
A detailed complexity analysis is provided in Appendix~\ref{append_t&s_complexity_analysis}.

\subsection{Faster \textit{Non-selfishness} Principle Graph Coarsening}

Although \texttt{NOPE} provides an effective coarsening strategy, the rapid growth of node degrees during merging incurs significant computational overhead due to the quadratic dependence of its time complexity on node degree. Under the assumption of local isomorphism, where adjacent node features exhibit minimal directional variation, we propose a faster variant, \texttt{NOPE}$^{*}$. This variant replaces the exact importance measure $\mathcal{I}$ with its expectation $\mathbb{E}[\mathcal{I}]$, achieving substantially improved computational efficiency.

\begin{algorithm}[h]
   \caption{\texttt{NOPE}}
   \label{alg:nope}
\begin{algorithmic}[1]  
   \STATE {\bfseries Input:} Graph $\mathcal{G}=(\mathcal{V}, \mathcal{E}, \mathcal{X})$, Merge ratio $r$
   \STATE {\bfseries Output:} Coarsened graph structure and embeddings

   \STATE Initialize $\mathbf{x}\in \mathbb{R}^{n_{max}\times d}$, $\mathbf{s}\in\mathbb{R}^{n_{max}}$, $\mathcal{A} \leftarrow \mathcal{V}$
   \STATE $\mathbf{x}[0:n] \leftarrow \mathcal{X}$, $\mathbf{s}[0:n]\leftarrow \mathbf{1}_n$
   \STATE Initialize sum-square cache $\boldsymbol{\Sigma} \in \mathbb{R}^{n_{max}}$:
   \STATE $\boldsymbol{\Sigma}[i] \leftarrow \sum_{k \in \mathcal{N}_i} (\mathbf{x}_k^\top \mathbf{x}_i)^2~~~\forall i \in \mathcal{V}$
   
    \STATE Initialize Min-Heap $\mathcal{H}$
    \STATE $\mathcal{H}.\text{push}((\mathcal{I}(u,v,\boldsymbol{\Sigma}), u, v))~~~\forall (u,v) \in \mathcal{E}$ 
   
   \WHILE{$|\mathcal{A}| > |\mathcal{V}| \cdot (1-r)$}
       \STATE $(\mathcal{I}_{uv}, u, v) \leftarrow \mathcal{H}.\text{pop}()$
       \IF{$u \notin \mathcal{A} \lor v \notin \mathcal{A}$} \STATE \textbf{continue}
       \ENDIF
       
       \STATE Create new node $w$ with index $|\mathcal{V}| + 1$
       \STATE Update node size: $\mathbf{s}_w \leftarrow \mathbf{s}_u + \mathbf{s}_v$
       \STATE Update embedding: $\mathbf{x}_w \leftarrow (\mathbf{s}_u \mathbf{x}_u + \mathbf{s}_v \mathbf{x}_v) / \mathbf{s}_w$
       \STATE Update neighbors: $\mathcal{N}_w \leftarrow (\mathcal{N}_u \cup \mathcal{N}_v) \setminus \{u, v\}$
       
       \IF{$\mathcal{N}_w \neq \emptyset$}
           \STATE Let $\mathbf{x}_{\mathcal{N}} \in \mathbb{R}^{|\mathcal{N}_w| \times D}$ be neighbor feature matrix
           \STATE $\mathbf{p}_w \leftarrow \langle\mathbf{x}_{\mathcal{N}},\mathbf{x}_w\rangle$, $\mathbf{p}_u \leftarrow \langle\mathbf{x}_{\mathcal{N}},\mathbf{x}_u\rangle$, $\mathbf{p}_v \leftarrow \langle\mathbf{x}_{\mathcal{N}},\mathbf{x}_v\rangle$
           \STATE $\boldsymbol{\Sigma}[w] \leftarrow \|\mathbf{p}_w\|_2^2$
           \STATE Masks $\mathbf{m}_u \in \{0,1\}^{|\mathcal{N}_w|}$ where $(\mathbf{m}_u)_k = \mathbf{1}_{k \in \mathcal{N}_u}$
           \STATE Masks $\mathbf{m}_v \in \{0,1\}^{|\mathcal{N}_w|}$ where $(\mathbf{m}_v)_k = \mathbf{1}_{k \in \mathcal{N}_v}$
           \STATE $\boldsymbol{\epsilon} \leftarrow \mathbf{p}_w^2 - (\mathbf{p}_u^2 \odot \mathbf{m}_u + \mathbf{p}_v^2 \odot \mathbf{m}_v)$
           \STATE $\boldsymbol{\Sigma}[\mathcal{N}_w] \leftarrow \boldsymbol{\Sigma}[\mathcal{N}_w] + \boldsymbol{\epsilon}$
       \ENDIF
       
       \STATE Update $\mathcal{A} \leftarrow (\mathcal{A} \setminus \{u, v\}) \cup \{w\}$
       \FOR{$k \in \mathcal{N}_w$}
           \STATE $\mathcal{H}.\text{push}(\text{Calculate}~\mathcal{I}(w, k, \boldsymbol{\Sigma}))$
           \STATE $\mathcal{N}_k\leftarrow\mathcal{N}_k\setminus \{u,v\}$, $\mathcal{N}_k\leftarrow\mathcal{N}_k \cup \{w\}$
       \ENDFOR
   \ENDWHILE
\end{algorithmic}
\end{algorithm}



\subsubsection{Expected Neighborhood Interference}
We first introduce the assumption of \emph{local isotropy}~\citep{node_lis, lis}.

\begin{assumption}[\textit{Local Isotropy}]\label{ass_1}
Conditioned on the reference node $r$, the neighborhood embeddings satisfy
\begin{equation*}
\mathbb{E}[\mathbf{x}_i] = 0,\quad 
\mathbb{E}[\mathbf{x}_i\mathbf{x}_i^T] = \sigma^2 \mathbf{I}_d,\quad \forall i \in \mathcal{N}_r,
\end{equation*}
where $\sigma^2>0$ denotes the variance associated with the reference node $r$ and $\mathbf{I}_d$ is the $d$-dimensional identity matrix.
\end{assumption}

Under Assumption~\ref{ass_1}, the expectation of $\mathcal{I}$ admits the following closed-form expression
\begin{equation}\label{expected_NI}
\mathbb{E}[\mathcal{I}] 
= \sum_{i\in\mathcal{N}_p\cup \mathcal{N}_q}
\frac{|v_p^c||v_q^c|}{|v_p^c|+|v_q^c|}
\sigma^2 \|\mathbf{x}_p-\mathbf{x}_q\|_2^2.
\end{equation}

We denote $\mathbb{E}[\mathcal{I}]$ by $\mathcal{I}^*$ for brevity, the proof is provided in Appendix~\ref{proof_e_NI}. 
Compared to Eq.~\ref{i_simple}, Eq.~\ref{expected_NI} aggregates neighbor contributions through a shared-neighbor weighting term, retaining neighborhood information while avoiding neighbor-wise computations.

\subsubsection{\texttt{NOPE}$^*$ Algorithm}

The pseudo-code of \texttt{NOPE}$^{*}$ is provided in Algorithm~\ref{alg:mem_eff_m2} (Appendix~\ref{append_pseucode_nope*}), Here we focus only on the key differences from Algorithm~\ref{alg:nope}. Compared with Algorithm~\ref{alg:nope}, Algorithm~\ref{alg:mem_eff_m2} replaces the exact measure $\mathcal{I}$ with its expectation $\mathcal{I}^*$, and updates the corresponding cache. This substitution collapses explicit neighbor-wise dot-product computations into a single degree-weighted feature difference term, enabling efficient incremental updates during coarsening.
As a result, the complexity of each merge evaluation is reduced from $O(\Delta\cdot d)$ to $O(d)$, leading to a substantial improvement in computational efficiency.

\subsubsection{\texttt{NOPE}$^*$ Time and Space Complexity Analysis}

As discussed above, \texttt{NOPE}$^{*}$ reduces the cost of updating $\mathcal{I}$ for a single merge to $\mathcal{O}(d)$. 
Over $\mathcal{O}(n)$ merge operations, the overall time complexity is therefore approximately $\mathcal{O}(n \cdot \Delta \cdot d)$. 
Hence, the dependence on node degree is reduced from quadratic to linear, significantly improving scalability. This allows the generated dense graph to remain tractable during the coarsening process and enables the algorithm to be applied to relatively extreme coarsening scenarios.

In terms of memory consumption, \texttt{NOPE}$^{*}$ maintains a linear space complexity of $\mathcal{O}(n_{\max} d + |\mathcal{E}|)$, without introducing intermediate neighbor-level buffers. Compared to \texttt{NOPE}, it reduces the peak memory consumption. 
A detailed complexity analysis is provided in Appendix~\ref{append_t&s_complexity_analysis_nope*}.

\subsubsection{Merge Process Analysis}

\begin{figure}[!h]
\begin{center}
\centerline{\includegraphics[width=0.85\columnwidth]{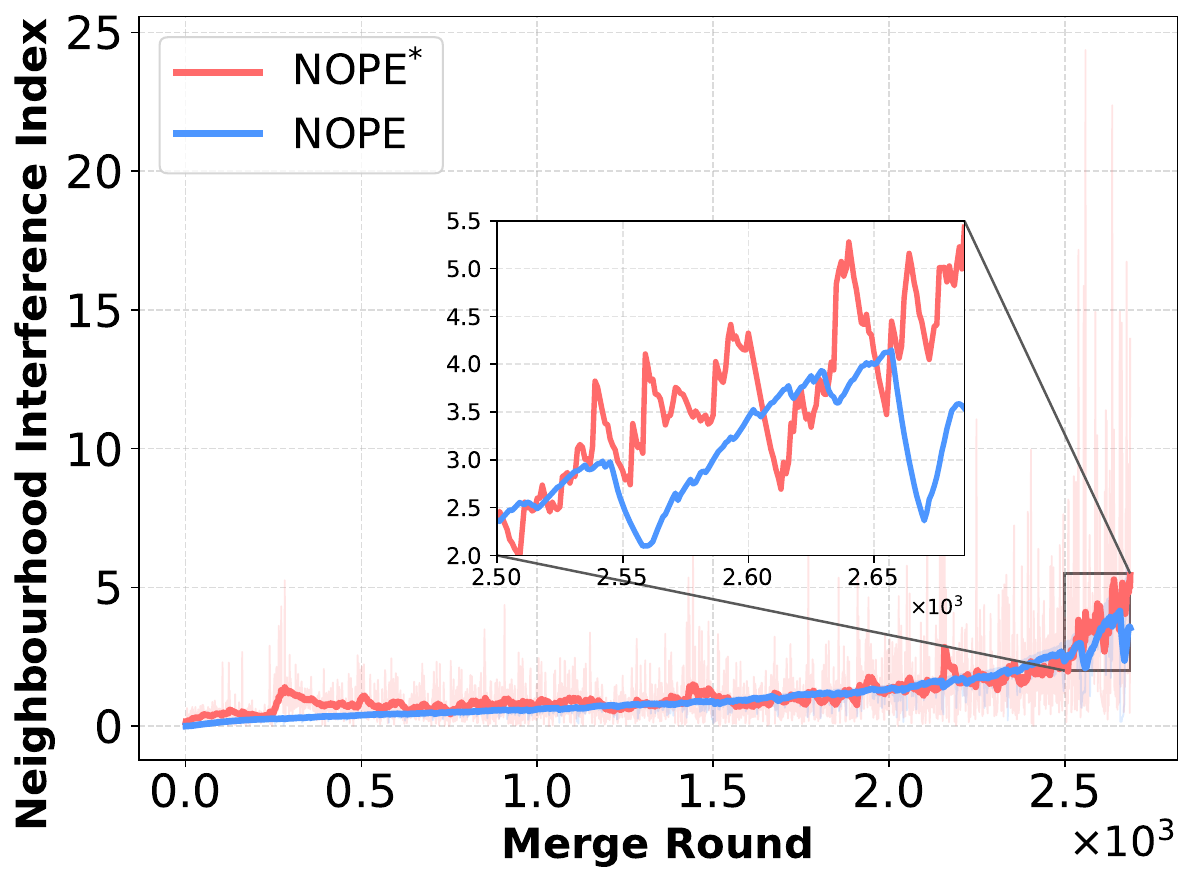}}
\caption{The neighborhood interference index $\mathcal{I}$ measured at each round of the \texttt{NOPE} and \texttt{NOPE}$^*$.}
\label{round_id}
\end{center}
\vspace{-6mm}
\end{figure}

In this section, we investigate how \texttt{NOPE} and its faster variant \texttt{NOPE}$^*$ behave along the coarsening trajectory. Figure~\ref{round_id} reports the neighborhood interference index $\mathcal{I}$ across merge rounds on Citeseer.


In the early stage, the EWMA~\cite{ewma} trends of $\mathcal{I}$ under \texttt{NOPE}$^*$ closely match those of \texttt{NOPE}, indicating that $\mathcal{I}^*$ provides an accurate surrogate for $\mathcal{I}$ when local isotropy (Assumption~\ref{ass_1}) holds .
As coarsening proceeds, the two curves gradually diverge and $\mathcal{I}$ under \texttt{NOPE}$^*$ increases more rapidly, suggesting growing approximation bias as merges become larger and more heterogeneous.


Taken together, \texttt{NOPE}$^*$ is most effective during the early-to-middle coarsening stages, where it achieves interference levels comparable to \texttt{NOPE}, or only mildly higher. As the merger progresses, \texttt{NOPE}$^*$ transitions into a higher-risk regime, characterized by increasing surrogate bias. In summary, our results suggest that applying \texttt{NOPE} at low to moderate coarsening rates is preferable.

\begin{table*}[ht]
\centering
\caption{Comparison of running memory and time on different datasets. Among them, OOM (out of memory) represents that the memory usage caused by the algorithm is greater than 96G, and OOT (out of time) indicates that the running time exceeds 5 hours.}
\resizebox{\textwidth}{!}{
\begin{tabular}{llcccccc}
\toprule
\multicolumn{2}{c}{\multirow{2}{*}{\textit{r= 0.5}}}
& \multicolumn{4}{c}{\textbf{Feature Graph Coarsening}} 
& \multicolumn{2}{c}{\textbf{Ours}} \\
\cmidrule(lr){3-6}
\cmidrule(lr){7-8}
\multicolumn{2}{c}{} & FGC & MPG & UGC & A-CM & \texttt{NOPE} & \texttt{NOPE}$^*$ \\
\midrule

\multirow{2}{*}{Citeseer} 
& Running Memory 
& 187.93MB & 471.11MB & 59.50MB & 96.72MB & 17.50MB & \textbf{16.05MB} \\
& Runtime
& 2113.41s & 89.07s & 1.30s & 4.09s & 0.97s & \textbf{0.35s} \\
\midrule

\multirow{2}{*}{Products(subset)} 
& Running Memory 
& / & / & 4331.60MB & 1228.96MB 
& \textbf{240.62MB} & 243.06MB \\
& Runtime 
& OOT & OOT & 3min1.76s & 1min9.38s 
& 11.57s & \textbf{6.46s} \\
\midrule

\multirow{2}{*}{Ogb-Arxiv} 
& Running Memory 
& OOM & OOM & 53,885.60MB & 6,526.96MB
& \textbf{1,307.60MB} & 1,354.61MB \\
& Runtime 
& / & / & 37min17.51s & 16min1.00s 
& 5min12.79s & \textbf{1min4.38s} \\
\midrule

\multirow{2}{*}{Book} 

& Running Memory 
& OOM & OOM & OOM &  27,351.26MB
& 5,542.74MB & \textbf{4,837.80MB} \\
& Runtime 
& / & / & / & 1h21min6.50s 
& 44min48.80s & \textbf{4min29.46s} \\
\midrule

\multirow{2}{*}{Products} 
& Running Memory 
& OOM & OOM & OOM & /
& / & \textbf{35,348.38MB} \\
& Runtime 
& / & / & / & OOT 
& OOT & \textbf{1h50min9.36s} \\
\bottomrule
\end{tabular}
}
\label{tab:efficiency}
\end{table*}

\begin{figure*}[!t]
\begin{center}
\centerline{\includegraphics[width=2\columnwidth]{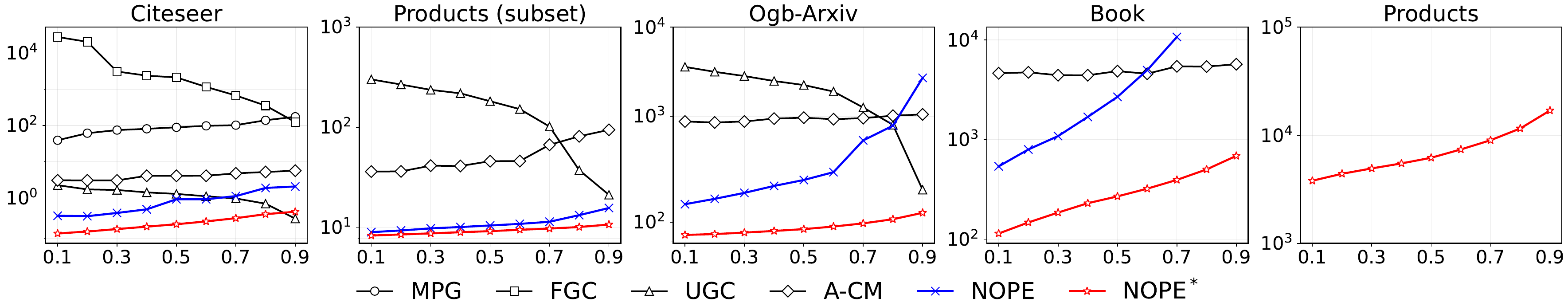}}
\caption{Time consumption in five datasets under different ratios $r$.}
\label{time_cost_r}
\end{center}
\end{figure*}


\begin{figure*}[!t]
\vspace{-5mm}
\begin{center}
\centerline{\includegraphics[width=2\columnwidth]{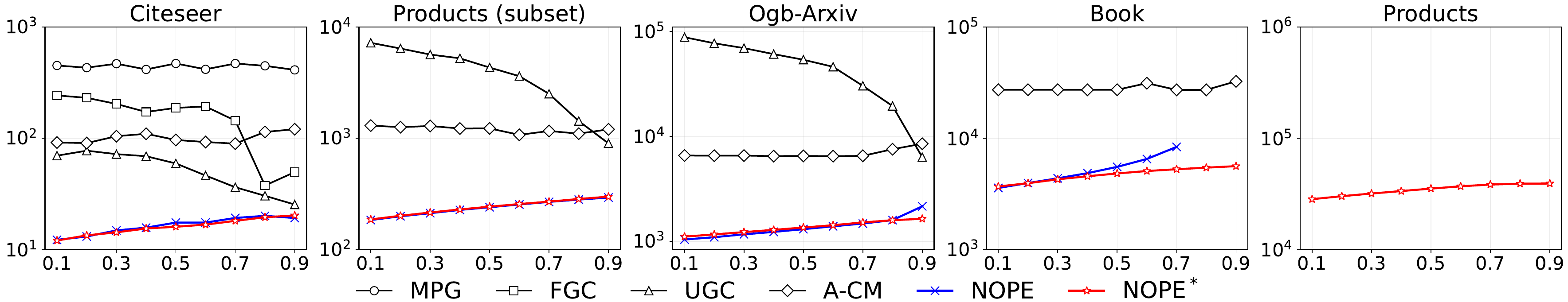}}
\caption{Memory consumption in five datasets under different ratios $r$.}
\label{mem_cost_r}
\end{center}
\vspace{-6mm}
\end{figure*}

\section{Experiments}
In this section, we first introduce the experimental settings, including the datasets, evaluation metrics, and baselines. Then, we evaluate our proposed \texttt{NOPE} and \texttt{NOPE}$^*$, with extensive experiments to answer the following research questions (RQs):
\begin{itemize}
    \item \textbf{RQ1}: How do \texttt{NOPE} and \texttt{NOPE}$^*$ compare to other baselines in terms of runtime and memory consumption?
    \item \textbf{RQ2}: In node classification tasks, how do the coarsened graphs generated by \texttt{NOPE} and \texttt{NOPE}$^*$ perform compared to those from other baseline methods?
    \item \textbf{RQ3}: How do the inference results on the coarsened graph generated by the \texttt{NOPE} and \texttt{NOPE}$^*$ models differ across different coarsening ratios?
\end{itemize}

\subsection{Experimental Settings}

\textbf{Datasets}. To evaluate the performance, we conduct node-level classification tasks on 6 real-world datasets, Citeseer~\cite{ogbn}, Ogb-Arxiv~\cite{ogbn_arxiv}, Book~\cite{book}, and Products~\citep{ogbn, products1, products2}. The detailed information is summarized in Appendix~\ref{datasets}.

\textbf{Baseline}. We compare the proposed \texttt{NOPE} and \texttt{NOPE}$^*$ with four feature graph coarsening algorithms, including two graph algorithms that guarantee coarsening performance via $\epsilon-similarity$~\cite{fgc_jmlr} theory: FGC~\cite{fgc_jmlr}, UGC~\cite{ugc_nips}, and two that seek to preserve downstream GNN performance: MPG~\cite{gc-mpg_nips}, A-CM~\cite{a-cm}. We also benchmarked performance against classical node classification models based on GNNs, including GCN~\cite{gcn}, GIN~\cite{gin} and GraphSAGE~\cite{graphsage}, while referencing conventional neighborhood sampling methods such as Random sampling~\cite{random,random2}, Degree sampling~\cite{graph_skeleton}, and RAG sampling~\cite{rag}. We aim for the algorithm's performance on coarsened graphs to approximate as closely as possible that of the original graph. For detailed model introduction and parameter settings, please refer to Appendix~\ref{baseline}.

\textbf{Evaluation Methods}. 
For the LLM-based node classification task, we select 10 nodes and employ LLMs for text reasoning to obtain corresponding text summaries for each supernode, i.e.
\vspace{-1.2mm}
\begin{equation*}
R_{v^c}
= f_{\mathrm{LLM}}\!\left(
\mathrm{Prompt}_{supernode}\!\left(
\{R_i \mid v_i \in \mathcal{S}_{v^c}\},\;
\theta
\right)
\right)
\vspace{-1.6mm}
\end{equation*}

where $\mathcal{S}_{v^c} \subseteq v^c,~\left| \mathcal{S}_{v^c} \right|= \min\!\left( |v^c|,\, 10 \right)$ and $\theta$ denotes the parameters of the LLM.
For single node label prediction, we jointly performed label prediction on both the target node's text and the text of its parent supernode, that is
\vspace{-1.3mm}
\begin{equation*}
    \hat{y}_i = f_{LLM}(\mathrm{Prompt}_{predict}(\{v_i,v^c|v_i\in v^c\}),\theta)
\end{equation*}
All inference tasks are conducted by Llama 3.1 8B-Instruct~\cite{dllama}.

For GNN-based node classification tasks, GNN message passing is first performed on the coarsened graph. Subsequently, predictions are made for target nodes using both their original features and the GNN-output features, i.e.:
\vspace{-1.9mm}
\begin{equation*}
    \hat{y} = \text{MLP}_{classifier}(\text{MLP}_{node}(\mathbf{x}_i)||\text{GNN}_{(\mathcal{G}^c, \mathcal{X}^c)}(v_i)).
\end{equation*}
Here $\text{MLP}_{node}$ denotes the node feature encoder, $\text{GNN}_{(\mathcal{G}^c,\mathcal{X}^c)}$ refers to the output of the GNN model on the coarsened graph $\mathcal{G}^c$, and $\text{MLP}_{classifier}$ indicates the node label predictor.
For specific prompt design and the implementation details, please refer to the Appendix~\ref{prompt_temp}.

\begin{table*}[!h]
\centering
\small
\setlength{\tabcolsep}{6pt}
\caption{LLM node classification results on different datasets under $r=0.5$.}
\renewcommand{\arraystretch}{1.15}
\scalebox{1}{
\begin{tabular}{ll|ccc|cccccc}
\toprule
\multicolumn{2}{c}{\textbf{Node Classifiction}} &
\multicolumn{3}{c}{\textit{Full Graph}} &
\multicolumn{4}{c}{\textbf{Graph Coarsening}} &
\multicolumn{2}{c}{\textbf{Ours}}\\
\cmidrule(lr){1-2}\cmidrule(lr){3-5}\cmidrule(lr){6-9}\cmidrule(lr){10-11}
\textbf{Dataset} & \textbf{Metric} &
\textit{Random} & \textit{Degree} & \textit{RAG} &
FGC & MPG & UGC & A-CM &
\texttt{NOPE} & \texttt{NOPE}$^*$ \\
\midrule

\multirow{2}{*}{Citeseer}
& ACC
& \textit{0.5924} & \textit{0.5768} & \textit{0.5956}
& 0.5911 & 0.6112 & 0.5893 & \underline{0.6081}
& \textbf{0.6206} & 0.6018 \\
& F1
& \textit{0.6098} & \textit{0.5972} & \textit{0.6147}
& 0.6038 & 0.6280 & 0.6049 & \underline{0.6221}
& \textbf{0.6403} & 0.6138 \\
\midrule

\multirow{2}{*}{Products(subset)}
& ACC
& \textit{0.5761} & \textit{0.5859} & \textit{0.5789}
& / & / & 0.5484 & 0.6755
& \underline{0.6783} & \textbf{0.6790} \\
& F1
& \textit{0.5980} & \textit{0.6078} & \textit{0.6003}
& / & / & 0.5676 & 0.6754
& \underline{0.6820} & \textbf{0.6827} \\
\midrule

\multirow{2}{*}{Ogb-Arxiv}
& ACC
& \textit{0.4029} & \textit{0.4142} & \textit{0.4256}
& / & / & 0.3482 & \textbf{0.3925}
& \underline{0.3808} & 0.3795 \\
& F1
& \textit{0.3940} & \textit{0.4046} & \textit{0.4154}
& / & / & 0.3456 & \textbf{0.3818}
& \underline{0.3738} & 0.3737 \\
\midrule

\multirow{2}{*}{Book}
& ACC
& \textit{0.8996} & \textit{0.8994} & \textit{0.9080}
& / & / & / & 0.9162
& \underline{0.9170} & \textbf{0.9244} \\
& F1
& / & / & /
& / & / & / & /
& / & / \\
\bottomrule
\end{tabular}
}
\label{tab:nc_llm_0.5}
\end{table*}

\begin{figure*}[!t]
\begin{center}
\centerline{\includegraphics[width=2\columnwidth]{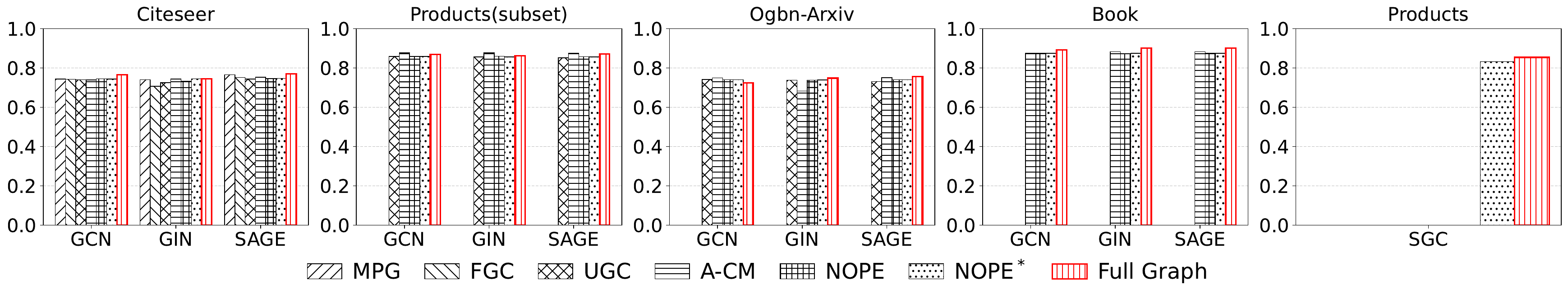}}
\caption{Accuracy/Hamming loss for GNN node classification in five datasets under $r= 0.5$.}
\label{nc_gnn_acc_0.5}
\end{center}
\vspace{-5mm}
\end{figure*}

\begin{figure*}[!t]
\begin{center}
\centerline{\includegraphics[width=2\columnwidth]{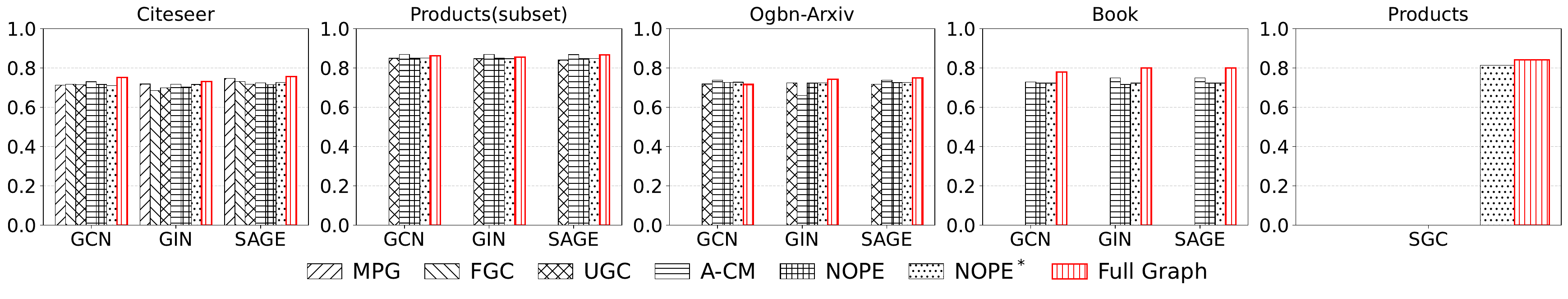}}
\caption{F1-score for GNN node classification in five datasets under $r= 0.5$.}
\label{nc_gnn_f1_0.5}
\end{center}
\vspace{-8mm}
\end{figure*}

\subsection{Runtime and Memory Consumption Analysis (RQ1)}

Table~\ref{tab:efficiency} compares runtime and memory consumption under a fixed coarsening ratio of 0.5. Overall, \texttt{NOPE} and \texttt{NOPE}$^*$ consistently outperform existing graph coarsening methods.

On the small Citeseer, all methods are feasible, yet other approaches incur substantial computational overhead. In contrast, \texttt{NOPE}$^*$ achieves a 0.35s runtime with only 16MB memory, yielding orders-of-magnitude speedups over FGC and MPG. As graph scale increases, the performance gap becomes more pronounced. On Products(subset) and Ogb-Arxiv, several baselines fail due to OOT or OOM, whereas \texttt{NOPE} and \texttt{NOPE}$^*$ complete within seconds to minutes using significantly less memory (e.g., 1.35GB vs. 53GB on Ogb-Arxiv). On large-scale datasets (Book and Products), most baselines become infeasible; notably, \texttt{NOPE}$^*$ is the only method that successfully processes the full Products graph under the 96GB memory constraint.

Figures~\ref{time_cost_r} and~\ref{mem_cost_r} further evaluate the time and memory efficiency under different coarsening ratios $r$. 
\texttt{NOPE} exhibits good time efficiency on most datasets at low and moderate coarsening ratios. 
However, as the coarsening rate increases, leading to degraded performance. Similar to the previous analysis, this effect arises from the significant increase in node degrees in the coarsened graph. In contrast, the faster variant \texttt{NOPE}$^{*}$ remains efficient and stable in all settings. In terms of memory consumption, \texttt{NOPE} and \texttt{NOPE}$^{*}$ consistently achieve the lowest memory usage across all datasets and coarsening ratios, demonstrating strong memory efficiency and stability.

\begin{figure*}[!t]
\begin{center}
\centerline{\includegraphics[width=2\columnwidth]{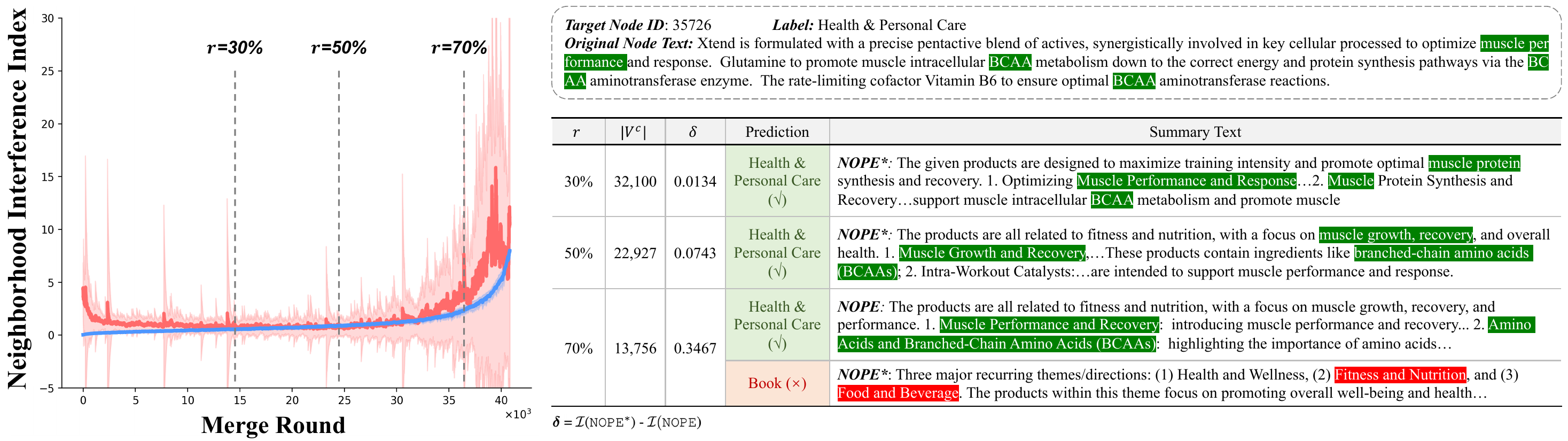}}
\caption{The case study on the Products dataset respectively presents the text of the supernode generated by the \texttt{NOPE} and \texttt{NOPE}$^*$ algorithms under different coarsening rates $r$, as well as the corresponding final prediction.}
\label{case_graph}
\end{center}
\vspace{-6mm}
\end{figure*}

\subsection{Node Classification Performance (RQ2)}

\subsubsection{LLM-based Node Classification}

Table~\ref{tab:nc_llm_0.5} summarizes the performance of different methods on LLM-based node classification under a coarsening ratio of 0.5. Overall, the performance of \texttt{NOPE} is significantly better than heuristic benchmark methods such as Random, Degree, and RAG. Even on the Products(subset), it achieves an increase of over 15\%. This highlights the clear advantages of node merging with interference awareness in terms of semantic richness and fidelity.
Compared with FGC and UGC, \texttt{NOPE} achieves improved results on all graphs, while \texttt{NOPE}$^*$ largely follows the same trend, suggesting that semantic information required by LLM-based classifiers can be effectively preserved even under lightweight approximations. In addition, compared to GNN-specific graph coarsening methods such as MPG and A-CM, \texttt{NOPE} and \texttt{NOPE}$^*$ exhibit more stable performance across different graphs. Notably, A-CM achieves marginal improvements only on Ogb-Arxiv, where heuristic neighborhood retrieval is particularly effective, partially accounting for the competitiveness of similarity-based coarsening on this graph.

Results under coarsening ratios of 0.3 and 0.7, reported in Tables~\ref{tab:nc_llm_0.3} and~\ref{tab:nc_llm_0.7}, exhibit consistent performance trends across methods. Notably, \texttt{NOPE} and \texttt{NOPE}$^*$ maintain competitive and stable performance under both settings, further demonstrating their robustness to varying coarsening rates.

\subsubsection{GNN-based Node Classification}

Results for GNN-based node classification under a coarsening ratio of 0.5 are reported in Figures~\ref{nc_gnn_acc_0.5} and~\ref{nc_gnn_f1_0.5}. Compared with full-graph GNN models, \texttt{NOPE} achieves performance that remains close to that obtained on the original graphs across most graphs, suggesting that the coarsened graphs retain a large portion of the information required for GNN inference.
Relative to feature-based graph coarsening methods such as FGC and UGC, \texttt{NOPE} shows comparable performance in most settings, indicating that interference-aware coarsening is a viable alternative under GNN-based evaluation. Moreover, although MPG and A-CM are strong GNN-specific baselines tailored to exploit structural properties, \texttt{NOPE}, while being model-agnostic, achieves broadly comparable results and in some cases performs favorably on certain graphs. The efficient variant \texttt{NOPE}$^*$ generally follows similar trends, with modest variations in performance.

Additional results under coarsening ratios of 0.3 and 0.7 are presented in Figures~\ref{nc_gnn_0.3_acc},~\ref{nc_gnn_0.3_f1} and Figures~\ref{nc_gnn_0.7_acc},~\ref{nc_gnn_0.7_f1}. Across different coarsening rates, the relative performance trends remain largely consistent, with \texttt{NOPE} and \texttt{NOPE}$^*$ exhibiting comparable behavior and alternating advantages across datasets, indicating stable performance under varying levels of graph reduction.

\subsection{Case Study (RQ3)}
In this section, we present a representative case to illustrate supernode text generation across different coarsening rates. Take a node (ID 35726, label: Health \& Personal Care) from the Products(subset) as a case, the supernode summaries generated by \texttt{NOPE} and \texttt{NOPE}$^*$, assessing their support for accurate label prediction.

Under low and moderate coarsening rates ($r=0.3$ and $r=0.5$), both \texttt{NOPE} and \texttt{NOPE}$^*$ produce semantically coherent summaries that align well with the target node, preserving key concepts related to muscle performance, recovery, and BCAAs. In these cases, the target category is correctly identified by both methods. At a high coarsening rate ($r=0.7$), however, the two methods diverge in this example. While \texttt{NOPE} continues to capture the core semantics of the target node, \texttt{NOPE}$^*$ yields less focused and less discriminative summaries, resulting in an incorrect prediction. This behavior is accompanied by increased neighborhood interference, suggesting that approximation errors under aggressive coarsening may introduce semantically heterogeneous neighbors.

Overall, this case study highlights a representative failure mode of \texttt{NOPE}$^*$ at high coarsening rates. Nevertheless, quantitative results show that such cases are uncommon, and \texttt{NOPE}$^*$ remains accurate in the majority of scenarios.




\section{Conclusion}
In this work, we introduce a \textit{non-selfishness} principle for graph coarsening that explicitly accounts for neighborhood-level interference during node aggregation. By formulating the neighborhood interference index ($\mathcal{I}$), we move beyond pairwise global similarity and enable neighborhood-aware merging that preserves local semantic consistency. Building on this principle, we propose \texttt{NOPE}, a greedy coarsening algorithm with near-linear time complexity and linear memory usage, along with its fast approximation \texttt{NOPE}$^*$, which reduces the overhead of interference computation at higher coarsening ratios. Extensive experiments on large text-attributed graphs show that \texttt{NOPE} and \texttt{NOPE}$^*$ achieve favorable efficiency–performance trade-offs for both LLM-based and GNN-based node classification, while significantly reducing runtime and memory consumption compared to existing methods.

Looking ahead, the interference-aware formulation provides a flexible foundation for future extensions, including dynamic or temporal graphs and scaling to billion-scale networks through streaming, distributed and so on.

\section*{Acknowledgments}

We would like to thank the reviewers for their suggestions
to improve this paper.  This work is supported by National Natural Science Foundation of China (No. T2421002, 62602003, 62272293) and Postdoctoral Fellowship Program of CPSF under Grant Number No. GZB20250806.

\section*{Impact Statement}
This paper presents work whose goal is to advance the field of Machine Learning. There are many potential societal consequences of our work, none which we feel must be specifically highlighted here.


\bibliography{main}
\bibliographystyle{icml2026}

\newpage
\appendix
\onecolumn

\section{Notations and Corresponding Meanings}
In this section, we supplement the paper with a detailed list of notations and their corresponding meanings, as listed in Table~\ref{tab:notation}.
\begin{table*}[]
    \centering
    \caption{Notations and corresponding meanings.}
    \begin{tabular}{cccc}
    \hline
        \textbf{Notation} &  \textbf{Meaning} &\textbf{Notation} &  \textbf{Meaning}\\
        \hline
        $\mathcal{G}$ & Text-attributed graph (TAG) & $\mathcal{V}$ & All nodes in TAG\\
        $\mathcal{E}$ & All edges in TAG & $\mathcal{R}$ & The raw text sequences of nodes\\
        $\mathcal{X}$ & The feature matrix of nodes & $n$ & The number of nodes in $\mathcal{G}$\\ 
        $d$ & The dimension of feature & $v_i$ & Node of $\mathcal{V}$\\
        $\mathcal{N}_{v_i}$ & The neighbor set of node $v_i$ & $\mathbf{x}_i$/$\mathcal{X}[v_i]$ & Feature vector of node $v_i$\\
        $\mathcal{G}_c$ & Coarsened text-attributed graph & $\mathcal{V}_c$ & All nodes in coarsened TAG\\ 
        $v^c_i$ & Node in $\mathcal{G}_c$ & $u/v/w$ & Node in $\mathcal{G}$ or $\mathcal{G}_c$ \\
        $\mathcal{E}_c$ & All edges in coarsened TAG & $|v^c_i|$ & Size of node set $v^c_i$\\ 
        $s_{ij}$ & Similarity between nodes $v_i$ and $v_j$ & $n_{max}$ & The number of pre-allocated nodes\\ 
        $\mathbf{C}$ & The mapping matrix & $r$ & The coarsening rate\\
        \text{degree} & The average degree of $\mathcal{G}$ & $\Delta$ & The dynamic average degree during coarsening\\
        \hline
    \end{tabular}
    \label{tab:notation}
\end{table*}

\section{Some Inference about $\mathcal{I}$.}\label{nni_sim_infer}
For convenience, we consider the case where supernodes $v_p^c$ and $v_q^c$ are merged to form node $v_w^c$.
\subsection{Embedding Update}\label{emb_update}

As seen in part \ref{prob_form} above, the embedding of node $v^c_w$ in the coarsened graph $\mathcal{G}_c$ is obtained by averaging the embeddings of its corresponding nodes in the original graph $\mathcal{G}$, i.e.,$\mathbf{x}^c_w = \frac{1}{|v_w^c|} \sum_{j \in v_w^c} \mathbf{x}_j$. Hence, the embedding updating process can be formulated as
\begin{equation*}
    \begin{split}
        \mathbf{x}_w^c&=\frac{1}{|v_w^c|} \sum_{i\in v_w^c} \mathbf{x}_w^c=\frac{1}{|v_p|+|v_q|}(\sum_{i \in v_p^c}\mathbf{x}_i + \sum_{j\in v_q^c}\mathbf{x}_j)=\frac{1}{|v_p^c|+|v_q^c|}(|v_p^c|\cdot\frac{1}{|v_p^c|}\sum_{i \in v_p^c}\mathbf{x}_i + |v_q^c|\cdot\frac{1}{|v_q^c|}\sum_{j \in v_q^c}\mathbf{x}_j)\\
                      &= \frac{|v_p^c|\mathbf{x}_p^c+|v_q^c|\mathbf{x}_q^c}{|v_p^c|+|v_q^c|}.
    \end{split}
\end{equation*}
\subsection{Simplification Process of Index $\mathcal{I}$}\label{simp_pro_i}
Given 
\begin{equation*}
    \mathcal{I}_{pq} = \sum_{i\in \mathcal{N}_{p}\cup \mathcal{N}_q} |v_p|(s_{ip}-s_{iw})^2 + |v_q|(s_{iq}-s_{iw})^2,
\end{equation*}
where
\begin{equation*}
    \begin{split}
        &s_{iw}=\langle \mathbf{x}_i, \mathbf{x}_r\rangle = \mathbf{x}_i^T \frac{|v_p^c|\mathbf{x}_p^c+|v_q^c|\mathbf{x}_q^c}{|v_p^c|+|v_q^c|} =  \frac{|v_p^c|s_{ip}+|v_q^c|s_{iq}}{|v_p^c|+|v_q^c|};\\
        &s_{ip}-s_{iw} = \frac{|v_p^c|s_{ip}-|v_q^c|s_{iq}}{|v_p^c|+|v_q^c|},~s_{iq}-s_{iw}=\frac{|v_p^c|s_{iq}-|v_q^c|s_{ip}}{|v_p^c|+|v_q^c|}.\\        
    \end{split}
\end{equation*}
Then
\begin{equation*}
    \begin{split}
        &\mathcal{I}_{pq} = \sum_{i\in \mathcal{N}_{p}\cup \mathcal{N}_q} (\frac{|v_p^c||v_q^c|^2}{(|v_p^c|+|v_q^c|)^2}+\frac{|v_p^c|^2|v_q^c|}{(|v_p^c|+|v_q^c|)^2})(s_{ip}-s_{iq})^2=\sum_{i\in \mathcal{N}_{p}\cup \mathcal{N}_q}\frac{|v_p^c||v_q^c|}{|v_p^c|+|v_q^c|}(s_{ip}-s_{iq})^2.
    \end{split}
\end{equation*}

\subsection{Experimental Design}\label{reasonable_exp_design}
To examine the impact of the \textit{non-selfishness} principle, we conduct controlled experiments comparing a \textit{non-selfishness} coarsening process with a \textit{selfishness} counterpart that selects merge pairs solely based on cosine similarity. Both of these methods employ the same greedy coarsening process, ensuring that any behavioral differences are solely due to the impact of whether a \textit{non-selfishness} strategy is adopted. The experiment is conducted on the Citeseer citation network. During each merging round, the Dirichlet Energy of the coarsened graph and the Average Edge Betweenness Centrality (Avg. EBC) are recorded.

(1) Dirichlet Energy: a semantic measure of feature smoothness on the graph, where higher values indicate stronger feature variation and better preservation of high-frequency semantic information. Its specific form is
\begin{equation*}
    \text{Dirichlet Energy} = \frac{1}{|\mathcal{E}|}\sum_{(i,j)\in\mathcal{E}}||\mathbf{x}_i-\mathbf{x}_j||_2^2.
\end{equation*}
(2) Average Edge Betweenness Centrality (Avg. EBC) is a structural measure that reflects how shortest-path connectivity is distributed across edges during coarsening, indicating whether connectivity responsibility remains evenly shared or becomes excessively concentrated on a small subset of critical edges, which may also manifest as premature degradation of community structure. It can be written as
\begin{equation*}
    \text{Avg EBC} = \frac{1}{|\mathcal{E}|} \sum_{e \in \mathcal{E}} \sum_{s,t \in \mathcal{V}, s < t} \frac{\sigma_{st}(e)}{\sigma_{st}},
\end{equation*}
where $\sigma_{st}$ represent the total number of shortest paths between nodes $s$ and $t$, and $\sigma_{st}(e)$ denotes the number of those shortest paths passing through edge $e$.

\section{Time and Space Complexity Analysis of \texttt{NOPE}}\label{append_t&s_complexity_analysis}

\textbf{Time Complexity}: Initializing the priority queue requires computing the index $\mathcal{I}$ for all edges. First, the cache items are pre-computed in $\mathcal{O}(n \cdot \text{degree} \cdot d)$, resulting in a total heap construction time of $\mathcal{O}(\mathcal{E}(\text{degree} \cdot d + \log \mathcal{E}))$. Under the assumption that $\mathcal{E} \approx \frac{1}{2}\cdot \mathcal{N} \cdot \text{degree}$, this complexity approximates to $\mathcal{O}(n \cdot \text{degree}^2 \cdot d)$. In the iterative phase, merging nodes generates approximately $\Delta$ new edges. For each new edge, computing $\mathcal{I}$ involves identifying common neighbors and calculating dot product differences, which incurs a cost of $\mathcal{O}(\Delta \cdot d)$. Consequently, the dominant cost per iteration arises from handling these new edges: $\Delta \cdot \mathcal{O}(\Delta \cdot d) = \mathcal{O}(\Delta^2 \cdot d)$.

Considering the algorithm performs $\mathcal{O}(n)$ merge operations, the overall complexity is approximate to  $\mathcal{O}(n\cdot\Delta^2\cdot d)$. In sparse graphs where $\Delta \ll n$, this approaches linearity with respect to graph size.

\textbf{Space Complexity}: 
The spatial requirements are primarily dictated by the storage of high-dimensional node features and the graph topology. In terms of feature storage, the algorithm utilizes a pre-allocation strategy to avoid memory fragmentation. The feature matrix vectors requires $\mathcal{O}(n_{max}d)$  space. Regarding graph topology, the adjacency list and the priority queue (min-heap) store graph connectivity, consuming $\mathcal{O}(|\mathcal{E}|)$ space. The auxiliary arrays for node sizes and the structural cache require $\mathcal{O}(n)$ space. Moreover, during the vectorized operations, temporary matrices of size $\mathcal{O}(\Delta\cdot d)$ are created, but these are transient and do not impact the asymptotic bound.

The overall space complexity is $\mathcal{O}(n_{max}d+|\mathcal{E}|)$. This is linear with respect to the input graph size, making the algorithm memory-efficient for large-scale, text-attributed graphs where $n$ and $d$ are large.

\section{Some Inference about \texttt{NOPE}$^*$ Algorithm}\label{infer_noper}
\subsection{Proof of the Expectation of $\mathcal{I}$ ($\mathbb{E}(\mathcal{I})$)}\label{proof_e_NI}
Let $\delta \triangleq \mathbf{x}_p-\mathbf{x}_q$, $\mathcal{N}_r=\mathcal{N}_p\cup\mathcal{N}_q$ and $M\triangleq\sum_{i\in \mathcal{N}_{p}\cup \mathcal{N}_q}\mathbf{x}_i\mathbf{x}_i^T\in\mathbb{R}^{d\times d}$, then the $\mathcal{I}_{pq}$ can be written as
\begin{equation*}
    \begin{split}
        \mathcal{I}_{pq} &= \sum_{i\in \mathcal{N}_{p}\cup \mathcal{N}_q}\frac{|v_p^c||v_q^c|}{|v_p^c|+|v_q^c|}(s_{ip}-s_{iq})^2= \sum_{i\in \mathcal{N}_{p}\cup \mathcal{N}_q}\frac{|v_p^c||v_q^c|}{|v_p^c|+|v_q^c|}((\mathbf{x}_i^c)^T(\mathbf{x}_p^c-\mathbf{x}_q^c))^2 = \frac{|v_p^c||v_q^c|}{|v_p^c|+|v_q^c|} \delta^T M \delta.
    \end{split}
\end{equation*}
Given that the assumption~\ref{ass_1}, the expectation of $\mathbb{E}[\mathcal{I}]$ is
\begin{equation*}
    \mathbb{E}[\mathcal{I}] = \frac{|v_p^c||v_q^c|}{|v_p^c|+|v_q^c|}\delta^T\mathbb{E}[M]\delta = \frac{|v_p^c||v_q^c|\sigma^2}{|v_p^c|+|v_q^c|}|\mathcal{N}_r|||\mathbf{x}_p-\mathbf{x}_q||_2^2.
\end{equation*}

\subsection{The Pseudocode of \texttt{NOPE}$^{*}$}~\label{append_pseucode_nope*}
The pseudocode of \texttt{NOPE}$^{*}$ can be found in Algorithm~\ref{alg:mem_eff_m2}.

\begin{algorithm}[tb]
   \caption{\texttt{NOPE}$^*$}
   \label{alg:mem_eff_m2}
\begin{algorithmic}[1]  
   \STATE {\bfseries Input:} Graph $\mathcal{G}=(\mathcal{V}, \mathcal{E}, \mathcal{X})$, Merge ratio $r$
   \STATE {\bfseries Output:} Coarsened graph structure and embeddings
   \STATE Initialize $\mathbf{x}\in \mathbb{R}^{n_{max}\times d}$, $\mathbf{s}\in\mathbb{R}^{n_{max}}$, $\mathbf{d}\in\mathbb{R}^{n_{max}}$, $\mathcal{A} \leftarrow \mathcal{V}$
   \STATE $\mathbf{x}[0:n] \leftarrow \mathcal{X}$, $\mathbf{s}[0:n]\leftarrow \mathbf{1}_n$, $\mathbf{d}[i] \leftarrow |\mathcal{N}_i|~~~\forall i \in \mathcal{V}$   
   \STATE Initialize Min-Heap $\mathcal{H}$
   \STATE $\mathcal{H}.\text{push}((\mathcal{I}^*(u,v), u, v))~~~\forall (u,v) \in \mathcal{E}$ 
   
   \WHILE{$|\mathcal{A}| > |\mathcal{V}| \cdot (1-r)$}
       \STATE $(\mathcal{I}^*(u,v), u, v) \leftarrow \mathcal{H}.\text{pop}()$
       \IF{$u \notin \mathcal{A} \lor v \notin \mathcal{A}$} \STATE \textbf{continue}
       \ENDIF
       
       \STATE Create new node $w$ with index $|\mathcal{V}| + 1$
       \STATE Update node size: $\mathbf{s}_w \leftarrow \mathbf{s}_u + \mathbf{s}_v$
       \STATE Update embedding: $\mathbf{x}_w \leftarrow (\mathbf{s}_u \mathbf{x}_u + \mathbf{s}_v \mathbf{x}_v) / \mathbf{s}_w$
       \STATE Update neighbors: $\mathcal{N}_w \leftarrow (\mathcal{N}_u \cup \mathcal{N}_v) \setminus \{u, v\}$
       \STATE Update degree: $\mathbf{d}[w] \leftarrow |\mathcal{N}_w|$
       
       \STATE Update $\mathcal{A} \leftarrow (\mathcal{A} \setminus \{u, v\}) \cup \{w\}$
       
       \FOR{$k \in \mathcal{N}_w$}
            \STATE $\mathcal{N}_k\leftarrow(\mathcal{N}_k\setminus \{u,v\})\cup \{w\}$
           \STATE Update neighbor degree $\mathbf{d}[k] \leftarrow |\mathcal{N}_k|$ 
           \STATE $w_{edge} \leftarrow (\mathbf{s}_w \cdot \mathbf{s}_k) / (\mathbf{s}_w + \mathbf{s}_k)$
           \STATE $d_{struct} \leftarrow \mathbf{d}[w] + \mathbf{d}[k] - |\mathcal{N}_w \cap \mathcal{N}_k| - 2$
           \STATE $d_{feat} \leftarrow \|\mathbf{x}_w - \mathbf{x}_k\|_2^2$
           
           \STATE $\mathcal{H}.\text{push}((w_{edge} \cdot w_{struct} \cdot w_{feat}), w, k))$
       \ENDFOR
   \ENDWHILE
\end{algorithmic}
\end{algorithm}

\subsection{Time and Space Complexity Analysis of \texttt{NOPE}$^*$}\label{append_t&s_complexity_analysis_nope*}

\textbf{Time Complexity}: The initialization phase involves computing $\mathcal{I}^*$ for all existing edges. and it is computed as the Euclidean distance with a cost of $\mathcal{O}( d)$. Consequently, constructing the initial min-heap takes $\mathcal{O}(\mathcal{E}(d+log\mathcal{E}))$, which is dominated by the feature distance calculation, resulting in $\mathcal{O}(\mathcal{E}d)\approx\mathcal{O}(n\cdot \text{degree}\cdot d)$. The core efficiency improvement of \texttt{NOPE}$^*$ lies in the iterative coarsening loop. In each iteration, merging nodes $u$ and $v$ into $w$ incurs a feature aggregation cost of $\mathcal{O}(d)$. Furthermore, retrieving the neighbors of node $w$ takes $\mathcal{O}(\Delta)$ time, and for each neighbor, they share the same neighborhood interference term, with the calculation time being $\mathcal{O}(d)$. Therefore, this part consumes a total of $\mathcal{O}(\Delta) + \mathcal{O}(d) \approx O(d)$ time. Thus, the complexity of each iteration is $\Delta\cdot\mathcal{O}(d)=\mathcal{O}(\Delta\cdot d)$. Considering that the algorithm performs $\mathcal{O}(n)$ merge operations, the overall time complexity is approximately $\mathcal{O}(n\cdot \Delta\cdot d)$. Therefore, the total time complexity of \texttt{NOPE}$^*$ is $\mathcal{O}(n\cdot \Delta\cdot d)$.

\textbf{Space Complexity}: 
The storage requirements of \texttt{NOPE}$^*$ mainly stem from the global feature matrix $\mathcal{X}_{all}\in \mathbb{R}^{n_{max}\times d}$. The graph topology (adjacency list and heap) requires $\mathcal{O}(\mathcal{E})$ space, while the auxiliary size and degree groups consume $\mathcal{O}(n_{max})$. Notably, compared with \texttt{NOPE}$^*$, it avoids the vectorized batch updating process, eliminating the creation of temporary neighbor matrices (which occupy $\mathcal{O}(\Delta\cdot d)$ in \texttt{NOPE}) during runtime. This ensures that the memory footprint remains linear, bounded by $\mathcal{O}(n_{max}d+|\mathcal{E}|)$, minimizing peak memory usage in resource-constrained environments.

\section{Datasets}\label{datasets}
\begin{table}[]
    \centering
    \caption{Detailed information and statistics of graphs.}
    \scalebox{1}{
    \begin{tabular}{c|ccccc}
    \hline
       Dataset  & Nodes & Edges & Avg degree & Class & Model\\
       \hline
       Citeseer & 3,327 & 4,732 & 2.84 & 6 & GNN,LLM\\
       Product-subset & 45,855 & 111,638 & 4.87 & 43 & GNN, LLM\\
       Ogb-Arxiv & 169,343 & 1,166,243 & 13.77 & 40 & GNN, LLM\\
       Book & 594,484 & 3,510,209 & 11.81 & 8 & GNN, LLM\\
       Products & 2,449,029 & 123,718,280 & 101.03 & 47 & GNN \\
       \hline
    \end{tabular}
    }
    \label{tab:graph_stat}
\end{table}

In this section, we introduce the detailed information of the experimental datasets. And we summarize the statistics in the Table~\ref{tab:graph_stat}.
\begin{itemize}
    \item \textbf{Citeseer}~\cite{ogbn} is a citation network comprising 3,186 scientific publications, where nodes represent individual papers and edges indicate citation relationships. Each node is associated with text attributes derived from the paper’s title and abstract. 
    \item \textbf{Ogb-Arxiv}~\cite{ogbn_arxiv} is a citation network comprising 169,343 Arxiv CS papers and their citation relationships from 40 different academic disciplines. Each node represents a paper, node text attribute is the title and abstract of paper, and each edge represents a citation relationship.
    \item \textbf{Product-subset}~\cite{ogbn_arxiv, products1, products2} is a co-purchase graph derived from the Amazon product network, where each node represents a product item and an edge indicates that two products are frequently co-purchased. We adopt the version provided by TAPE~\cite{tape}, which is a subset of the original OGBN-Products dataset~\cite{products1}, comprising 45,855 nodes and 111,638 edges. Each node is associated with text attributes such as the product title and description. 
    \item \textbf{Book}~\cite{book} is a literary network from GoodReads, where each node represents a book and the edges represent their similarity relationships. It includes a total of 594,484 books. The node text feature represents the title and description of the book.
    \item \textbf{Products} is the original dataset of Product-subset, which includes 2,449,029 nodes and 123,718,280 edges. The node features of each node are composed of the first 100 final features obtained by PCA.
\end{itemize}
For all datasets, 10\% of the nodes are selected as the testing set. Additionally, in GNN node classification experiments, an extra 50\% of the nodes are selected as the training set and 10\% as the validation set.

\section{Baseline}\label{baseline}
Baseline model:
\begin{itemize}
    \item \textbf{FGC} proposes an optimization-driven graph coarsening framework that simultaneously utilizes both the graph structure and node features. By jointly optimizing the coarsened graph and its feature, it theoretically guarantees similarity between the coarsened graph and the original graph within the $\epsilon\in [0,1)$ range while preserving key graph properties. For the model parameters, where regularization parameters $\lambda=500, \beta=0$, relaxation parameters $\alpha=500, \gamma=384$.
    \item \textbf{UGC} proposes the Universal Graph Coarsening (UGC) method, which jointly models node attributes and adjacency structures while incorporating dataset dissimilarity factors to adaptively process homophonic and heterophonic graphs. By ensuring spectral similarity and $\epsilon-similarity$, UGC achieves efficient graph reduction while significantly enhancing downstream task performance and computational efficiency. For the model parameters. we select 1,000 hash projectors and each initialized with a uniform norm within the interval [0,1].
    \item \textbf{MPG} proposal a specialized message passing mechanism tailored for coarsened graphs by redesigning the propagation operator on coarsened graphs to enable directed propagation even when the original graph is undirected, while theoretically ensuring signal propagation integrity. In this paper, we set the maximum number of nodes merged at one coarsening step $n_e$ as 100.
    
    \item \textbf{A-CM} proposes ConvMatch, a graph aggregation method based on convolutional matching, along with its efficient variant A-ConvMatch. By matching and preserving graph convolutional outputs, it directly aligns the convolutional operations of graph neural networks during the aggregation phase, achieving extreme-scale graph coarsening. Regarding parameter selection, we adhere to the original paper's settings by merging only the top-1 node pairs per round. While this approach impacts model efficiency, it enables precise control over coarsening ratios and delivers optimal results.
\end{itemize}
Reference model:
\begin{itemize}
    \item \textbf{Random}: For each target node, a neighbor node is randomly selected from its one-hop neighborhood without considering any structural or attribute-related criteria. Subsequently, the neighbor's text is concatenated with the central node's text attributes for evaluation.
    \item \textbf{Degree}: For each target node, select the neighbor node with the highest degree from its one-hop neighborhood. Subsequently, concatenate the neighbor's text with the central node's text attributes for evaluation.
    \item \textbf{RAG}: For each target node, retrieve the semantically most relevant neighbor node from its one-hop neighborhood. Subsequently, concatenate the neighbor text with the central node's textual attributes for evaluation.
    \item \textbf{GCN}: GCN achieves feature propagation and fusion based on graph structure by performing normalized weighted summation of neighbor node features. It leverages Laplacian smoothing to learn node representations, making it well-suited for processing graph data with strong homophony.
    \item \textbf{GIN}: GIN employs learnable summation aggregation functions combined with multilayer perceptrons (MLPs), theoretically possessing discriminative capabilities equivalent to the Weisfeiler–Lehman test. It emphasizes structural discriminability and is suitable for modeling fine-grained structures.
    \item \textbf{GraphSAGE}: GraphSAGE supports inductive learning by sampling neighbors and generating node representations using aggregation functions (mean, pooling, LSTM, etc.), enabling efficient inference on unseen new nodes or new graphs.
    \item \textbf{SGC}: SGC decouples the linear propagation from the nonlinear components in multi-layer GCN, eliminating intermediate nonlinearities and weight matrices. It retains only multi-step feature propagation and linear classifiers, significantly reducing computational complexity while achieving performance comparable to GCN.
\end{itemize}

\section{Supplementary Prompt Template}\label{prompt_temp}

\subsection{Prompt Design}

\begin{promptbox}{The Prompt Template for Supernode Summarization Text}
\textbf{Input}: Given a group of \texttt{\detokenize{{NODE_TYPE}}}s from the same community within a \texttt{\detokenize{{GRAPH_TYPE}}} graph. Each \texttt{\detokenize{{NODE_TYPE}}} is represented by its \texttt{\detokenize{{INFO}}}.

\textbf{Task}: Summarize the content of these \texttt{\detokenize{{NODE_TYPE}}}s that are semantically aligned with each other. The summary must explicitly articulate the intersection of their work.

Format requirement: (1) Identify the major recurring themes/directions; (2) Structure the summary around these identified themes; (3) Write the final summary in a cohesive and formal academic style.

\textbf{Answer}:
\end{promptbox}

\begin{promptbox}{The Prompt Template for Target Node Label Prediction}
\textbf{Input}: Given a \texttt{\detokenize{{GRAPH_TYPE}}} graph, the target \texttt{\detokenize{{NODE_TYPE}}} has the following information: \texttt{\detokenize{{TARGET_RAW_TEXT}}}. The target \texttt{\detokenize{{NODE_TYPE}}} is related to the following \texttt{\detokenize{{NODE_TYPE}}} summary: \texttt{\detokenize{{SUMMARY_TEXTS}}}

\textbf{Task}: Based on the features of the target \texttt{\detokenize{{NODE_TYPE}}} and the \texttt{\detokenize{{NODE_TYPE}}} summary, please determine the most appropriate \texttt{\detokenize{{GRAPH_TYPE}}} sub-category for the target \texttt{\detokenize{{NODE_TYPE}}}.

Categories: \texttt{\detokenize{{CATEGORY_LIST}}}.

Please think about the categorization of the target \texttt{\detokenize{{NODE_TYPE}}} in a structured manner, and only output the single most relevant category of the target \texttt{\detokenize{{NODE_TYPE}}}. Do not give any reasoning or extra text for your answer.

\textbf{Answer}:

\end{promptbox}

\subsection{Implementation Details}

All algorithms are run on a Linux server running Ubuntu 20.04 LTS. All coarsening algorithms are executed on an Intel(R) Xeon(R) Gold 5117 @ 2.0GHz 14C28T (up to 24 cores). Additionally, all evaluation processes are conducted on a GPU (NVIDIA GeForce RTX 4090, 24GB memory) using CUDA 12.6.

\section{Model Performance under Different Coarsening Ratio $r$}\label{appendix_model_performance}

\begin{figure*}[!h]
\begin{center}
\centerline{\includegraphics[width=\columnwidth]{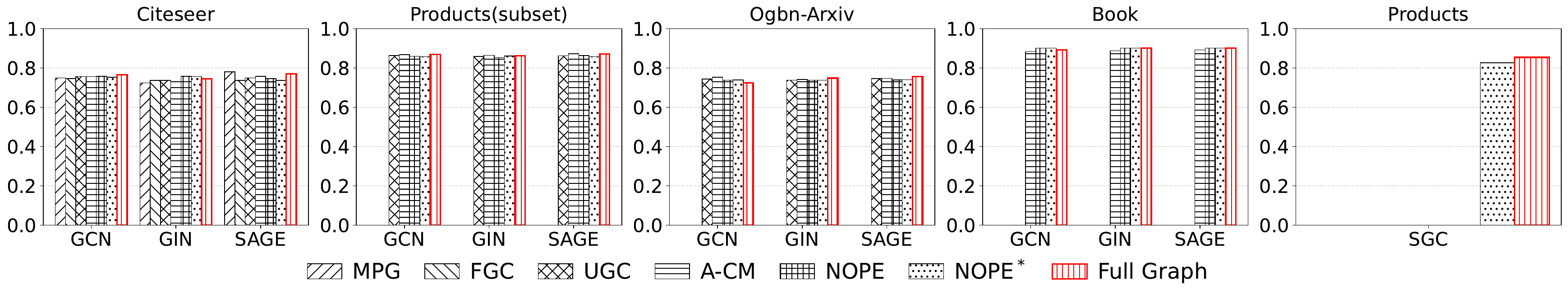}}
\caption{Accuracy/Hamming loss for GNN node classification in five datasets under $r= 0.3$.}
\label{nc_gnn_0.3_acc}
\end{center}
\vspace{-2mm}
\end{figure*}

\begin{figure*}[!h]
\begin{center}
\centerline{\includegraphics[width=\columnwidth]{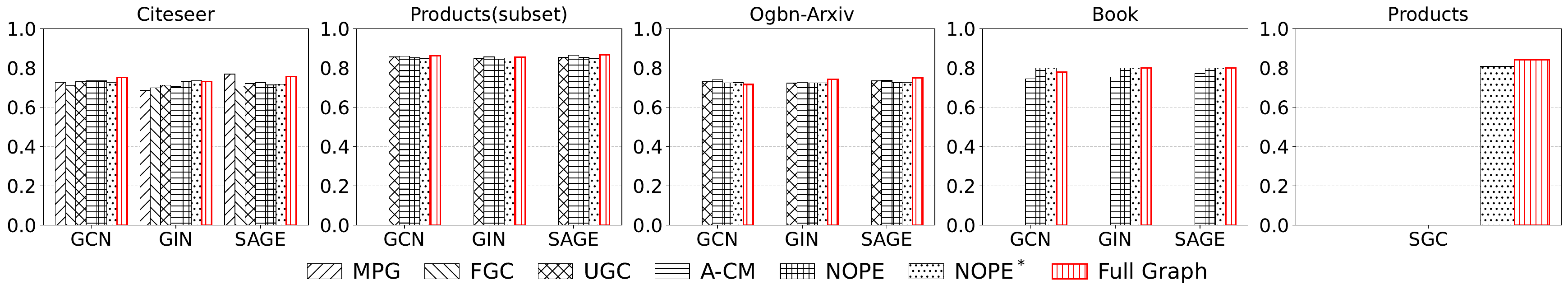}}
\caption{F1-score for GNN node classification in five datasets under $r= 0.3$}
\label{nc_gnn_0.3_f1}
\end{center}
\vspace{-2mm}
\end{figure*}

\begin{figure*}[!h]
\begin{center}
\centerline{\includegraphics[width=\columnwidth]{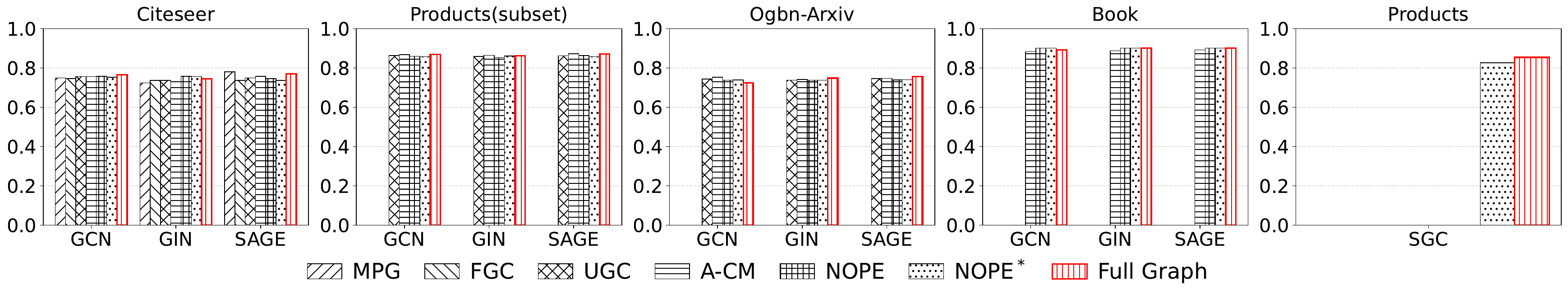}}
\caption{Accuracy/Hamming loss for GNN node classification in five datasets under $r= 0.7$.}
\label{nc_gnn_0.7_acc}
\end{center}
\vspace{-2mm}
\end{figure*}

\begin{figure*}[!h]
\begin{center}
\centerline{\includegraphics[width=\columnwidth]{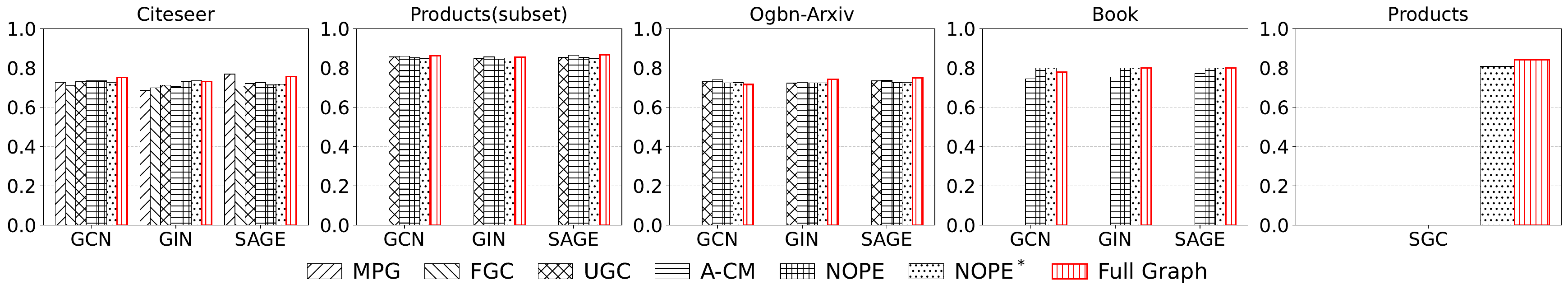}}
\caption{F1-score for GNN node classification in five datasets under $r= 0.7$.}
\label{nc_gnn_0.7_f1}
\end{center}
\vspace{-2mm}
\end{figure*}

In following Tables~\ref{sp_data_time_r} and~\ref{sp_data_mem_r}, to facilitate arrangement, we represent Products(subset) as Products$^*$ and abbreviate Ogb-Arxiv as Arxiv.

\begin{table*}[!ht]
\centering
\caption{The runtime of five datasets under different coarsening ratios.}
\label{sp_data_time_r}
\scalebox{0.75}{
\begin{tabular}{c|ccccccccc}
\hline
\hline
Citeseer & 0.1      & 0.2        & 0.3         & 0.4         & 0.5         & 0.6         & 0.7         & 0.8         & 0.9         \\
\hline
MPG      & 39.18s   & 1m1.30s  & 1m14.66s  & 1m20.95s  & 1m29.07s  & 1m37.71s  & 1m42.17s  & 2m19.27s  & 2m54.26s  \\
FGC      & OOT      & OOT        & 51m16.78s & 39m23.28s & 35m13.41s & 19m17.19s & 11m11.73s & 5m53.68s  & 2m3.97s   \\
UGC      & 2.26s    & 1.73s      & 1.67s       & 1.43s       & 1.30s       & 1.12s       & 0.99s       & 0.86s       & 0.49s       \\
A-CM     & 3.08s    & 3.07s      & 3.07s       & 4.10s       & 4.10s       & 4.12s       & 4.79s       & 5.23s       & 5.68s       \\
\texttt{NOPE}    & 0.56s    & 0.55s      & 0.63s       & 0.72s       & 0.97s       & 0.97s       & 1.13s       & 1.90s       & 2.08s       \\
\texttt{NOPE}$^*$    & 0.12s    & 0.17s      & 0.23s       & 0.29s       & 0.35s       & 0.42s       & 0.50s       & 0.60s       & 0.66s    \\
\hline
\hline
Products$^*$ & 0.1       & 0.2       & 0.3       & 0.4       & 0.5       & 0.6       & 0.7       & 0.8       & 0.9       \\
\hline
MPG      & /         & /         & /         & /         & /         & /         & /         & /         & /         \\
FGC      & /         & /         & /         & /         & /         & /         & /         & /         & /         \\
UGC      & 5m0.35s & 4m26.99s& 3m56.20s& 3m37.92s& 3m1.76s & 2m31.49s& 1m41.46s& 1m1.31s & 39.17s    \\
A-CM     & 1m0.07s & 1m0.23s & 1m5.35s & 1min5.08s & 1m9.41s & 1m9.52s & 1m24.12s& 1m31.79s& 1m37.62s\\
\texttt{NOPE}     & 5.54s     & 7.10s     & 8.96s     & 10.16s    & 11.57s    & 13.00s    & 14.91s    & 20.87s    & 27.28s    \\
\texttt{NOPE}$^*$    & 2.52s     & 3.48s     & 4.45s     & 5.49s     & 6.46s     & 7.71s     & 8.76s     & 10.17s    & 12.43s  \\
\hline
\hline
Arxiv   & 0.1          & 0.2          & 0.3          & 0.4          & 0.5          & 0.6          & 0.7          & 0.8          & 0.9          \\
\hline
MPG     & /            & /            & /            & /            & /            & /            & /            & /            & /            \\
FGC     & /            & /            & /            & /            & /            & /            & /            & /            & /            \\
UGC     & 59m37.36s  & 52m29.46s  & 47m2.28s   & 41m22.79s  & 37m17.51s  & 31m30.07s  & 20m50.89s  & 13m19.27s  & 4m23.16s   \\
A-CM    & 14m30.72s  & 14m9.38s   & 14m31.37s  & 15m39.30s  & 16m1.64s   & 15m24.22s  & 15m50.81s  & 16m52.37s  & 17m26.59s  \\
\texttt{NOPE}    & 3m10.36s   & 3m37.49s   & 4m7.68s    & 4m42.78s   & 5m12.79s   & 5m52.44s   & 8m54.41s   & 13m1.17s   & 44m52.75s  \\
\texttt{NOPE}$^*$   & 35.04s       & 39.44s       & 46.80s       & 55.24s       & 1m4.38s    & 1m17.51s   & 1m33.45s   & 1m54.46s   & 2m26.83s \\
\hline
\hline

Book    & 0.1          & 0.2          & 0.3          & 0.4          & 0.5          & 0.6          & 0.7           & 0.8          & 0.9          \\
\hline
MPG     & /            & /            & /            & /            & /            & /            & /             & /            & /            \\
FGC     & /            & /            & /            & /            & /            & /            & /             & /            & /            \\
UGC     & /            & /            & /            & /            & /            & /            & /             & /            & /            \\
A-CM    & 1h17m11.20s& 1h18m52.25s& 1h13m50.22s& 1h13m44.59s& 1h21m6.50s & 1h16m21.49s& 1h30m35.33s & 1h30m7.56s & 1h35m3.35s \\
\texttt{NOPE}    & 8m59.36s   & 13min16.56s  & 18m5.30s  & 28m11.31s & 44m48.80s & 1h22m51.96s& 2h58m32.87s & OOT          & OOT          \\
\texttt{NOPE}$^*$   & 1m54.29s   & 2m27.58s   & 3m5.46s   & 3m50.16s  & 4m29.46s  & 5m21.88s  & 6m34.47s    & 8m22.49s  & 11m27.32s \\
\hline
\hline
Products & 0.1            & 0.2            & 0.3            & 0.4            & 0.5            & 0.6            & 0.7            & 0.8            & 0.9            \\
\hline
MPG          & /              & /              & /              & /              & /              & /              & /              & /              & /              \\
FGC          & /              & /              & /              & /              & /              & /              & /              & /              & /              \\
UGC          & /              & /              & /              & /              & /              & /              & /              & /              & /              \\
A-CM         & /              & /              & /              & /              & /              & /              & /              & /              & /              \\
\texttt{NOPE}         & /              & /              & /              & /              & /              & /              & /              & /              & /              \\
\texttt{NOPE}$^*$       & 4h42m25.54s  & 3h12m36.13s  & 2h29m52.64s  & 2h3m12.45s   & 1h43m9.36s   & 1h31m27.73s  & 1h22m12.21s  & 1h13m21.48s  & 1h3m15.73s   \\
\hline
\hline

\end{tabular}
}
\end{table*}

\begin{table*}[!ht]
\caption{The consuming memory of five datasets under different coarsening ratio.}
\label{sp_data_mem_r}
\centering
\scalebox{0.74}{
\begin{tabular}{cccccccccc}
\hline
\hline
Citeseer & 0.1       & 0.2       & 0.3       & 0.4       & 0.5       & 0.6       & 0.7       & 0.8       & 0.9       \\
\hline
MPG      & 451.79MB  & 431.79MB  & 468.84MB  & 416.36MB  & 471.11MB  & 417.31MB  & 470.61MB  & 448.91MB  & 412.48MB  \\
FGC      & 243.12MB  & 231.53MB  & 203.83MB  & 172.72MB  & 187.93MB  & 193.47MB  & 144.30MB  & 37.67MB   & 49.74MB   \\
UGC      & 69.90MB   & 77.50MB   & 72.14MB   & 69.13MB   & 59.50MB   & 46.38MB   & 36.38MB   & 30.45MB   & 25.43MB   \\
A-CM     & 91.73MB   & 90.67MB   & 104.58MB  & 109.95MB  & 96.73MB   & 92.68MB   & 89.66MB   & 114.24MB  & 120.79MB  \\
\texttt{NOPE}     & 12.25MB   & 13.12MB   & 14.88MB   & 15.75MB   & 17.50MB   & 17.50MB   & 19.25MB   & 20.12MB   & 19.25MB   \\
\texttt{NOPE}$^*$    & 12.07MB   & 13.43MB   & 14.30MB   & 15.49MB   & 16.05MB   & 16.78MB   & 18.14MB   & 19.55MB   & 20.28MB   \\
\hline
\hline

Products$^*$ & 0.1        & 0.2        & 0.3        & 0.4        & 0.5        & 0.6        & 0.7        & 0.8        & 0.9        \\
\hline
MPG      & /          & /          & /          & /          & /          & /          & /          & /          & /          \\
FGC      & /          & /          & /          & /          & /          & /          & /          & /          & /          \\
UGC      & 7,225.50MB  & 6,421.10MB  & 5,666.80MB  & 5,241.80MB  & 4,331.60MB  & 3,623.00MB  & 2,513.60MB  & 1,431.00MB  & 901.50MB   \\
A-CM     & 1,302.47MB  & 1,261.29MB  & 1,291.98MB  & 1,225.85MB  & 1,228.96MB  & 1,075.68MB  & 1,164.57MB  & 1,102.46MB  & 1,205.55MB  \\
\texttt{NOPE}     & 184.62MB   & 199.50MB   & 212.62MB   & 227.50MB   & 240.62MB   & 254.62MB   & 268.62MB   & 281.75MB   & 294.00MB   \\
\texttt{NOPE}$^*$    & 299.14MB   & 285.18MB   & 269.91MB   & 256.79MB   & 243.06MB   &  229.73MB & 215.54MB & 201.39MB & 186.79MB\\
\hline
\hline
Arxiv   & 0.1         & 0.2         & 0.3         & 0.4         & 0.5         & 0.6         & 0.7         & 0.8         & 0.9         \\
\hline
MPG     & /           & /           & /           & /           & /           & /           & /           & /           & /           \\
FGC     & /           & /           & /           & /           & /           & /           & /           & /           & /           \\
UGC     & 88,335.60MB  & 77,347.90MB  & 69,565.90MB  & 60,981.60MB  & 53,885.60MB  & 46,148.70MB  & 30,388.40MB  & 19,513.00MB  & 6,346.90MB   \\
A-CM    & 6,580.30MB   & 6,553.71MB   & 6,572.69MB   & 6,498.28MB   & 6,526.97MB   & 6,488.29MB   & 6,536.53MB   & 7,555.96MB   & 8,528.61MB   \\
\texttt{NOPE}    & 1,041.51MB   & 1,095.64MB   & 1,167.07MB   & 1,230.91MB   & 1,307.60MB   & 1,391.21MB   & 1,479.27MB   & 1,595.02MB   & 2,156.32MB   \\
\texttt{NOPE}$^*$   & 1,109.54MB   & 1,164.48MB   & 1,225.16MB   & 1,287.05MB   & 1,354.61MB   & 1,425.54MB   & 1,515.77MB   & 1,590.35MB   & 1,642.17MB   \\
\hline
\hline
Book    & 0.1         & 0.2         & 0.3         & 0.4         & 0.5         & 0.6         & 0.7         & 0.8         & 0.9         \\
\hline
MPG     & /           & /           & /           & /           & /           & /           & /           & /           & /           \\
FGC     & /           & /           & /           & /           & /           & /           & /           & /           & /           \\
UGC     & /           & /           & /           & /           & /           & /           & /           & /           & /           \\
A-CM    & 27,333.19MB  & 27,357.56MB  & 27,362.19MB  & 27,290.74MB  & 27,351.27MB  & 31,317.80MB  & 27,276.52MB  & 27,283.93MB  & 32,553.43MB  \\
\texttt{NOPE}    & 3,581.41MB   & 3,969.21MB   & 4,374.80MB   & 4,867.13MB   & 5,542.74MB   & 6,529.45MB   & 8,376.42MB   & /           & /           \\
\texttt{NOPE}$^*$   & 3,695.39MB   & 3,952.46MB   & 4,284.29MB   & 4,563.70MB   & 4,837.80MB   & 5,087.12MB   & 5,279.05MB   & 5,453.23MB   & 5,628.05MB   \\
\hline
\hline
Citeseer & 0.1         & 0.2         & 0.3         & 0.4         & 0.5         & 0.6         & 0.7         & 0.8         & 0.9         \\
\hline
MPG      & /           & /           & /           & /           & /           & /           & /           & /           & /           \\
FGC      & /           & /           & /           & /           & /           & /           & /           & /           & /           \\
UGC      & /           & /           & /           & /           & /           & /           & /           & /           & /           \\
A-CM     & /           & /           & /           & /           & /           & /           & /           & /           & /           \\
\texttt{NOPE}     & /           & /           & /           & /           & /           & /           & /           & /           & /           \\
\texttt{NOPE}$^*$    & 39,288.35MB  & 39,131.04MB  & 38,455.35MB  & 37,070.87MB  & 35,348.83MB  & 33,603.52MB  & 31,926.63MB  & 30,209.79MB  & 28,408.45MB  \\
\hline
\hline
\end{tabular}
}
\end{table*}

\newpage

\begin{table*}[!h]
\centering
\caption{LLM node classification results on different datasets under $r=0.3$.}
\small
\setlength{\tabcolsep}{6pt}
\renewcommand{\arraystretch}{1.15}
\scalebox{1}{
\begin{tabular}{ll|ccc|cccccc}
\toprule
\multicolumn{2}{c}{\textbf{Node Classifiction}} &
\multicolumn{3}{c}{\textit{Full Graph}} &
\multicolumn{4}{c}{\textbf{Graph Coarsening}} & \multicolumn{2}{c}{\textbf{Ours}}\\
\cmidrule(lr){1-2}\cmidrule(lr){3-5}\cmidrule(lr){6-9}\cmidrule(lr){10-11}
\textbf{Dataset} & \textbf{Metric} &
\textit{Random} & \textit{Degree} & \textit{RAG} &
FGC & MPG & UGC & A-CM &
\texttt{NOPE} & \texttt{NOPE}$^*$ \\
\midrule

\multirow{2}{*}{Citeseer}
& ACC
& \textit{0.5768} & \textit{0.5924} & \textit{0.5956}
& 0.5924 & 0.6081 & 0.5799 & 0.6050 & 0.6207 & 0.5956 \\
& F1
& \textit{0.5972} & \textit{0.6098} & \textit{0.6147}
& 0.6093 & 0.6247 & 0.5944 & 0.6195 & 0.6366 & 0.6107 \\
\midrule

\multirow{2}{*}{Products(subset)}
& ACC
& \textit{0.5859} & \textit{0.5761} & \textit{0.5789}
& / & / & 0.6037 & 0.6690 & 0.6737 & 0.6796 \\
& F1
& \textit{0.6078} & \textit{0.5980} & \textit{0.6003}
& / & / & 0.6156 & 0.6698 & 0.6779 & 0.6830 \\
\midrule

\multirow{2}{*}{Ogb-Arxiv}
& ACC
& \textit{0.4142} & \textit{0.4029} & \textit{0.4256}
& / & / & 0.3475 & 0.3829 & 0.3746 & 0.3700 \\
& F1
& \textit{0.4046} & \textit{0.3940} & \textit{0.4154}
& / & / & 0.3432 & 0.3829 & 0.3746 & 0.3653 \\
\midrule

\multirow{2}{*}{Book}
& ACC
& \textit{0.8994} & \textit{0.8996} & \textit{0.9080}
& / & / & / & 0.9184 & 0.9200 & 0.9240 \\
& F1
& / & / & /
& / & / & / & / & / & / \\
\bottomrule
\end{tabular}
}
\label{tab:nc_llm_0.3}
\end{table*}

\begin{table*}[!h]
\centering
\caption{LLM node classification results on different datasets under $r=0.7$.}
\small
\setlength{\tabcolsep}{6pt}
\renewcommand{\arraystretch}{1.15}
\scalebox{1}{
\begin{tabular}{ll|ccc|cccccc}
\toprule
\multicolumn{2}{c}{\textbf{Node Classifiction}} &
\multicolumn{3}{c}{\textit{Full Graph}} &
\multicolumn{4}{c}{\textbf{Graph Coarsening}} &
\multicolumn{2}{c}{\textbf{Ours}}\\
\cmidrule(lr){1-2}\cmidrule(lr){3-5}\cmidrule(lr){6-9}\cmidrule(lr){10-11}
\textbf{Dataset} & \textbf{Metric} &
\textit{Random} & \textit{Degree} & \textit{RAG} &
FGC & MPG & UGC & A-CM & \texttt{NOPE} & \texttt{NOPE}$^*$ \\
\midrule

\multirow{2}{*}{Citeseer}
& ACC
& \textit{0.5768} & \textit{0.5924} & \textit{0.5956}
& 0.5817 & 0.6050 & 0.5517 & 0.6269 & 0.5987 & 0.6175 \\
& F1
& \textit{0.5972} & \textit{0.6098} & \textit{0.6147}
& 0.5963 & 0.6222 & 0.5711 & 0.6396 & 0.6150 & 0.6326 \\
\midrule

\multirow{2}{*}{Products(subset)}
& ACC
& \textit{0.5859} & \textit{0.5761} & \textit{0.5789}
& / & / & 0.5187 & 0.6768 & 0.6842 & 0.6881 \\
& F1
& \textit{0.6078} & \textit{0.5980} & \textit{0.6003}
& / & / & 0.5420 & 0.6780 & 0.6860 & 0.6917 \\
\midrule

\multirow{2}{*}{Ogb-Arxiv}
& ACC
& \textit{0.4142} & \textit{0.4029} & \textit{0.4256}
& / & / & 0.3200 & 0.4105 & 0.3880 & 0.3937 \\
& F1
& \textit{0.4046} & \textit{0.3940} & \textit{0.4154}
& / & / & 0.3121 & 0.3965 & 0.3785 & 0.3862 \\
\midrule

\multirow{2}{*}{Book}
& ACC
& \textit{0.8994} & \textit{0.8996} & \textit{0.9080}
& / & / & / & 0.9140 & 0.9104 & 0.9255 \\
& F1
& / & / & /
& / & / & / & / & / & / \\
\bottomrule
\end{tabular}
}
\label{tab:nc_llm_0.7}
\end{table*}


\end{document}